# Transparent Model of Unabridged Data (TMUD)

Jie Xu and Min Ding*

May 10, 2021

* Jie Xu is a Ph.D. candidate at School of Management, Fudan University, China (ines.xu@foxmail.com). Min Ding is Bard Professor of Marketing, Smeal College of Business, and Affiliate Professor, College of Information Sciences and Technology, Pennsylvania State University, USA (email: minding@psu.edu).

# Abstract

Recent advancements in computational power and algorithms have enabled unabridged data (e.g., raw images/audio) to be used as input in some models (e.g., deep learning). However, the black box nature of such models reduces their likelihood of adoption by marketing scholars. Our paradigm of analysis, the Transparent Model of Unabridged Data (TMUD), enables researchers to investigate the inner workings of such black box models by incorporating an *ex ante* filtration module and an *ex post* experimentation module. We empirically demonstrate the TMUD by investigating the role of facial components and sexual dimorphism in face perceptions, which have implications for four marketing contexts: advertisement (perceptions of approachability, trustworthiness, and competence), brand (perceptions of whether a face represents a brand's typical customer), category (perceptions of whether a face represents a category's typical customer), and customer persona (perceptions of whether a face represents the persona of a brand's customer segment). Our results reveal new and useful findings that enrich the existing literature on face perception, most of which is based on abridged attributes (e.g., width of mouth). The TMUD has great potential to be a useful paradigm for generating theoretical insights and may encourage more marketing researchers and practitioners to use unabridged data.

## Introduction

Approximately 80% of data used in marketing practice is unstructured (Rizkallah 2017) and contains rich information (e.g., visual and audio data) (Chintagunta et al. 2016, Urban et al. 2020, Xiao et al. 2013). Visual (e.g., YouTube, Instagram) and audio platforms (e.g., Spotify, Clubhouse) have acquired hundreds of millions or even billions of users. Consumers also actively generate and share a huge number of images, videos, and audio files on social media. Currently, nearly 1,000 photos are posted on Instagram each second, and more than 50 billion photos have been posted to date (Aslam 2021). As a result, many marketing scholars have begun to address important marketing problems using either image (e.g., Burnap et al. 2020, Dew et al. 2019, Dzyabura and Peres 2021, Dzyabura et al. 2021, Guan et al. 2020, Hartmann et al. 2021, Li and Xie 2020, Li et al. 2019, Liu et al. 2017b, Liu et al. 2020, Malik et al. 2019, Peng et al. 2020, Shin et al. 2020, Troncoso and Luo 2020, van der Lans et al. 2021, Xiao and Ding 2014, Zhang et al. 2014, Zhang et al. 2017, Zhang et al. 2021, Zhou et al. 2020) or video (e.g., Liu et al. 2018, Lu et al. 2016, Lu et al. 2020 , McDuff and Berger 2019, Teixeira et al. 2012, Teixeira et al. 2014, Tellis et al. 2019 , Tkachenko and Jedidi 2020, Zhang et al. 2020) data.

Unfortunately, these unstructured and rich data have posed a dilemma for researchers and practitioners. The traditional statistical approach requires extracting a set of so-called engineered features (independent variables) from unstructured data, but a lot of information is lost in the process. Alternatively, one can employ rapidly advancing deep learning models that are capable of accepting these rich and unstructured data to generate meaningful managerial insights, such as predictive tasks. Unlike the traditional statistical approach, however, deep learning models operate as black boxes (Urban et al. 2020, Xia et al. 2019, Hartmann et al. 2021, Gabel and Timoshenko 2021), as they are not designed to test theoretical insights. This problem is compounded by the sheer number of parameters that typically number in the millions or more. As a result, use of deep learning models has been limited, especially in marketing research aimed at generating theoretical insights. Even if generating theoretical insights is not the goal, the inability to understand how they work reduces confidence and trust in these models, further limiting the adoption of otherwise fruitful applications. As a result, few publications in marketing journals report deep learning models as the primary methodology (Burnap et al. 2020, Dew et al. 2019, Dzyabura et al., 2021, Gabel and Timoshenko 2021, Gabel et al. 2019, Guan et al. 2020, Hartmann et al. 2021, Hu et al. 2019, Li et al. 2019, Liu et al. 2019 , Liu et al. 2020, Malik et al. 2019, Shin et al. 2020, Timoshenko and Hauser 2019, Tkachenko and Jedidi 2020, Troncoso and Luo 2020, Xia et al. 2019, Zhang and Luo 2018, Zhang et al. 2021, Zhang et al., 2017)

Here, we propose the transparent model of unabridged data (TMUD), which resolves this dilemma by integrating several existing analytic tools. A model of unabridged data (MUD) uses raw data as input which is typically rich and unstructured. The word "transparent" reflects how a typical black box MUD can now be examined to understand its inner workings, thereby enabling theoretical investigation. The TMUD adds two modules to a typical black box model: an *ex ante* filtration module and an *ex post* experimentation module (Figure 1). The *ex ante* filtration and *ex post* experimentation modules do not need to be used simultaneously in TMUD, but combining the two will likely yield deeper insights into the inner workings of the black box model.

Insert Figure 1 here

The intuition for the *ex post* experimentation module comes from observing how humans learn and how researchers uncover their inference mechanisms after learning. The human mind can take raw input, the most important being visual and audio information, which is unstructured and rich, and make meaningful inferences and decisions accordingly. The human mind is unquestionably the most mysterious black box of all—it is extremely powerful, and yet we have only a rudimentary understanding of how it operates. Because human beings either are not aware of how they make decisions or do not want to disclose such information to outsiders, researchers have adopted scientific strategies to "open" the black box of the human mind. In disciplines that study human (or animal) behavior, researchers conduct well-designed controlled experiments in which only one or a few factors vary so they can uncover the theoretical underpinnings (i.e., inner workings) of specific human preferences and decisions. Following this established logic used to investigate humans' black box decision units—namely, how the human mind makes certain inferences and decisions—we propose adding the *ex post* experimentation module, a formalized process to understand the inner workings of machines' black box decision units (such as deep learning models).

The *ex post* experimentation module is activated after a black box is trained (i.e., learns from data), thus the adjective *ex post*. This module investigates how a trained black box forms preferences and makes decisions, without imposing any pre-conceived constraints such as functional form, variables to be included, etc. By experimentation, we refer to rigorous and scientifically designed experiments used in *ex post* studies. For example, controlled experiments may use inputs that only vary on one or a few chosen features (explanatory variables) of certain magnitude, which are otherwise identical (thereby controlling for other information). Like experiments designed to investigate the human mind, the *ex post* controlled experiment in TMUD feeds carefully designed input into a trained black box (such as a deep learning model), and then infers the effect of this explanatory variable (or mediator) by comparing the outputs of the black box from these otherwise identical inputs.

The *ex ante* filtration module is activated before the black box is trained, thus the word *ex ante*. The intuition for *ex ante* filtration is straightforward. If we have a mixture with 10 different elements that we know will cause a certain reaction (in a human or a machine) but we do not know which element is responsible, the original mixture can be filtered and individual elements can be tested to identify which one induces the reaction. More formally, filtration refers to signal filtering, a major domain of digital signal processing research in electronic engineering, where signals include audio, image, and video information, which also happen to be the overwhelming source of rich data used in black box models in marketing contexts. The idea of signal filtering is to apply a specific type of filter to a signal that allows only desirable elements to pass, thus creating a well-defined dataset for additional analysis. We incorporate this idea into TMUD by proposing that appropriate filters be applied to unabridged data to select only specific parts of the data, retaining full information about the focal aspects and eliminating other information. This filtered data (in a sense, purified data) is what is analyzed by the black box model. Since the input is clean, we can identify the role of a specific type of information in driving the inferences in the black box, which is nearly impossible when unfiltered data are used. This is conceptually similar to the compositional approach to preference measurement in marketing, such as the self-explicated method.

This paper makes two major contributions: a methodological contribution (i.e., the TMUD paradigm), and a theoretical contribution regarding the role of facial components and sexual dimorphism in face perception and implications for four important marketing contexts. First, to our knowledge, TMUD is the first systematic paradigm that can be used to investigate the inner workings of deep learning models (and black box models in general), identify how they operate, and test theoretical insights. Our innovation, which combines existing research in an unprecedented way for this specific purpose, will not only provide researchers with a tool to examine the inner workings of black box models, but also remove the stigma typically associated with deep learning models and thus encourage researchers to use—and we hope, even embrace—black box models in both research and applications in the future. Second, we demonstrate how a researcher/practitioner can use existing tools as building blocks to construct a TMUD to infer people's preferences for visual stimuli. In addition to demonstrating and validating how the TMUD can be used in real contexts, our empirical study also contributes substantively to literature on the role of facial components and sexual dimorphism in face perception in situations where such perception is of critical importance to marketing managers.

The rest of this paper is organized as follows. First, we describe the TMUD paradigm, focusing primarily on visual data. Then, we apply the TMUD to study facial components and sexual dimorphism on face perception across four major marketing contexts where such perceptions are critical: advertisement (perceptions of approachability, trustworthiness, and competence), brand (perceptions of whether a face represents a brand's typical customer), category (perceptions of whether a face represents a typical customer of the product category), and customer persona (perceptions of whether a face represents the persona of a brand's customer segment). The empirical application is presented in three separate sections to make it easier to follow. The paper concludes with a general discussion of the TMUD, including directions for future research.

## The TMUD Paradigm

In this section, we first provide definitions of the types of data (abridged vs. unabridged, phenomenon vs. component) and types of models (model of abridged data [MAD] vs. model of unabridged data [MUD], biological vs. machine) that are relevant to our investigation. We then examine the biological MUD and build intuition on how it works, especially how a third party can infer the underlying rules of the human mind, which is a black box to outsiders unless the human is willing and able to comprehensively articulate them. Based on this intuition, we propose the *ex post* experimentation module. Then, we propose the *ex ante* filtration module based on literature in electronic engineering on signal processing. After a brief comparison to the interpretability research in deep learning, we conclude this section by describing the value of the TMUD and its relevance to current research inquiries. Although we mainly explain and demonstrate our proposed TMUD in the context of visual data, our paradigm can be applied to other rich and unstructured data.

### Data: Abridged and Unabridged; Phenomenon and Component

We separate data along two dimensions: unabridged vs. abridged and phenomenon vs. component. These data may take many forms: visual, audio, smell, taste, touch, written language, complete behavioral data (e.g., everything a person did), and different combinations thereof. We use a print ad as an example below to illustrate these four different types of data. See Table 1 for additional examples of visual, audio, smell and taste data.

Insert Table 1 here

Merriam-Webster's dictionary defines a phenomenon as "a fact or event of scientific interest susceptible to scientific description and explanation." Based on this definition, a print ad is a phenomenon. A component of a phenomenon is self-contained—in other words, it contains all relevant information about itself and does not contain any information unrelated to itself. For example, the spokesperson in the print ad is a component, and the mouth in an image of a person's face is a component. A phenomenon may have one or many components.

Unabridged data constitute the complete set of information about a particular phenomenon of interest (i.e., all raw stimuli). In this example, the image of a print ad is unabridged data. The complete set of information related to a component of the phenomenon is also considered unabridged data if all information about the component is included and no unrelated information is included. In this example, a cutout of the spokesperson in the print ad is unbridged data as well. Unabridged data typically are rich in information and unstructured, but are not necessarily big data. For example, a picture is not big data. Abridged data constitute a set of abstracted features of unbridged data, either at the phenomenon level or component level, and typically are structured. For example, a print ad can be coded along several dimensions, such as the presence of people, props (e.g., a sofa), a color theme, an implied story (category variables), and so on. This coded profile of a print ad is abridged data. Similarly, we can obtain abridged data at the component level, such as the demographics of people represented in the data. In this paper, we use the phrases abstracted features and engineered features (a term from the deep learning literature) interchangeably.

Models: MAD vs. MUD; Biological vs. Machine

We group models along two dimensions (Table 2) based on whether they use abridged data or unabridged data, and whether they are biological models (the human mind) or machine models (including any statistical, machine learning, and deep learning models).

Insert Table 2 here

A model of abridged data (MAD) uses abridged data regarding a phenomenon (or one or more components) to make inferences about it. The functional form linking independent variables to dependent variables may (statistical model) or may not (deep learning model) be specified. A model of unabridged data (MUD) uses unabridged data regarding a phenomenon (or one or more components) to make inferences about it. It does not specify any functional forms, nor does it select *a priori* which features are important enough to be included. Typically, a MUD is considered a black box model since we do not know how it works, even if it works well in predictive tasks. A biological model is an internal mental model constructed by biological beings (such as humans) to understand a particular phenomenon. A machine model is externally constructed, typically using computers, to understand a particular phenomenon. For example, a human may identify that someone is sad by observing their facial expression (i.e., using the biological MUD) or by observing that they are crying (i.e., using the biological MAD). A convolutional neural network (CNN) classifier that can detect gender from a face image is a machine MUD, a model that predicts a person's gender based on the presence/absence of makeup is a machine MAD. Table 2 provides additional examples for these different types of models.

Before we perform an in-depth examination of the black box of the MUD, it is appropriate to examine the MAD which is typically used in marketing research, and contrast it with the existing MUD. Figure 2 provides a graphic comparison of how these two types of models operate. For the typical MAD, the original unabridged data with a known label (i.e., dependent variable) is first processed to generate a set of values representing *a priori* selected features (engineered features), which are structured. These data are then fed into a MAD to identify the best possible relationship between the set of engineered features and the label. Through this process, the MAD is able to identify the specific contribution of each engineered feature on the label based on its coefficient value. A user who wants to use the trained MAD to predict this label for a new phenomenon simply must extract the engineered features and combine them with the identified coefficients to make a prediction.

Insert Figure 2 here

In comparison, the MUD follows a much simpler process. It is trained directly by the raw unabridged data with the known label, which can be used to predict the label for a new phenomenon. On the other hand, it does not provide any insights regarding how it predicts a given label from the data. Thus, the MUD is considered a black box model.

Intuition from Inferences Made by the Biological MUD

We first examine the biological MUD and use it as the inspiration and template for building a transparent machine MUD. Biological beings have evolved over a long period of time to acquire the ability to make inferences about external stimuli and corresponding decisions (Figure 3). These external stimuli acquired by various sensory organs (i.e., visual, audio, smell, taste, and tactile information) typically are processed as unabridged data in the brains of biological beings. The biological MUD learns from these unabridged data and their associated labels (e.g., a potential predator, an angry human), and eventually acquires the ability to predict whether these labels should be applied to new phenomena.

Insert Figure 3 here

All biological MUDs are black box models since we do not know how biological minds work. Even many humans are not aware of how their own minds make inferences and decisions. If a third party wants to uncover the inference rules in a biological MUD, they typically rely on the experimental approach, either formally or informally.

Psychological research is illustrative in this regard. Psychology researchers are interested in how humans make inferences and decisions without asking them to explicitly describe the process (which they may not be able or willing to do). In a typical psychological research project, scholars create two or more stimuli that vary only on one specific variable. They expose different groups of randomly-selected human participants to different stimuli and solicit or observe their reactions. If their reactions differ, researchers are able to identify that the manipulated variable is an important factor in how humans make inferences and decisions. It is through such systematic investigation that researchers can tease out the theoretical insights that have driven a wide range of human inferences and behavior.

The *Ex Post* Experimentation Module

Building on the practice of how researchers obtain theoretical insights from a biological MUD, we formally describe the corresponding process used to obtain theoretical insights from a machine MUD (Figure 4)—namely, the *ex post* experimentation module. This process consists of two stages. The data generation stage involves creating the modified unabridged data, varying one or a few dimensions while controlling for all other dimensions. The analysis stage involves using the trained MUD to explore the effects of modified unabridged data on the predictive results and making appropriate interpretations.

Insert Figure 4 here

Two different approaches to modification can be taken. The first approach is direct and objective. For example, one may change the width of the mouth (a component) in the unabridged data of face (a phenomenon) to explore whether—and if so, how—the width of the mouth affects perceived trustworthiness. This approach does not involve investigating insights holistically, but is based on examining engineered features. On the other hand, it is relatively easy to implement, as the manipulated variable (e.g., width of the mouth) can often be adjusted with simple coding. We will not discuss this further in this paper. The second approach is indirect and often perceptual and holistic. For example, one may change faces to make them look more feminine, and explore whether, and if so, how perceived femininity affects perceived trustworthiness. This approach can generate substantially more interesting (holistic) theoretical insights, but is much harder to implement. This requires another set of models that can modify an existing phenomenon *holistically* along one specific perceptual dimension. Fortunately, recent advancements in deep learning have made this possible. Below, we describe how this process works with visual data.

Some of the most notable advancements are StyleGAN (Karras et al. 2019) and StyleGAN2 (Karras et al. 2020) which can generate artificial images that are indistinguishable from real images of objects (e.g., human faces, automobiles). StyleGAN and StyleGAN2 accomplish this feat by introducing a latent space consisting of 512 dimensions. Every object in a specific category is represented by a specific vector in this space. With sufficient training using real objects (e.g., faces), the models are able to generate artificial objects from latent space vectors that are indistinguishable from real objects. These models are agnostic about the categories of images used for training, and can be used to recreate any category of images as long as enough training images exist.

To the delight of researchers, it turns out that specific directions can be identified in the latent space, each corresponding to a unique feature (e.g., gender, eyeglasses, age, smile) (Radford et al. 2015, Shen et al. 2020). Radford et al.(2015) first demonstrated the arithmetic properties of a latent space vector, e.g., latent space vector (smiling woman) - latent space vector (neutral woman) + latent space vector (neutral man) = latent space vector (smiling man). For any binary semantic feature (e.g., male or female), a hyperplane can be identified as the separation boundary in the latent space. Editing features of images along the latent space vector which is orthogonal to the hyperplane ensures that images will vary on this specific binary semantic feature while other characteristics remain the same (Shen et al. 2020). For example, if we move along the latent space direction of male/female, a StyleGAN/StyleGAN2 trained on faces will gradually edit the image so that it becomes more feminine/masculine while still resembling the original face. This is exactly what is required to implement the second approach of modification.

This data generation approach involves three general steps. First, researchers need to encode the original data (e.g., face) in the latent space (i.e., identify the latent space vector corresponding to each face). Alternatively, these original faces could be randomly generated by the StyleGAN/StyleGAN2 based on their known latent space vector parameters. Second, the latent space direction corresponding to the feature of theoretical interest must be identified. Finally, faces are edited along this dimension to generate new and modified unabridged data (faces) where only one feature differs from the original unabridged data.

Regardless of the approach used to generate modified unabridged data, the effect of modifying a specific feature, as either explanatory variables or mediators, is tested in the analysis stage by systematically comparing the predictive outcomes of the MUD. At this stage, researchers can test any possible functional forms between the label (dependent variable) and the modified feature.

The *Ex Ante* Filtration Module

The rich data we are interested in (e.g., audio files, images, videos) are considered special cases of digital signals. A large literature (Lyons 2004, Oppenheim 1999) in electronic engineering focuses on digital signal processing, and one aim is to take a complex signal and isolate and/or understand the function of one or more of its specific components. This is often achieved through a filtering process that involves using various audio filters and image/video filters to eliminate unwanted information from the original signal and testing the "purified" signal to identify its roles. For example, a complex audio wave can be decomposed into its element frequencies through Fourier transformation prior to further analysis.

Following this literature, we propose to incorporate an *ex ante* filtration module that filters phenomenon level data to retain only information about a specific component. As defined earlier, a component is a part of a phenomenon that is self-contained—it contains all information about itself and does not contain any information unrelated to itself. By running the black box model using filtered data components one at a time, it is possible to estimate each component's contribution to the phenomenon level inference. If the component chosen is sufficiently small and meaningful, the analysis should yield rigorous theoretical insights and practical findings.

This process also makes it possible to investigate potential non-compensatory rules, especially disjunctive rules used in a MUD. For example, consider two components, *a* and *b*, in a phenomenon *T* that can be inferred on label *X*, where the sum of the predictive performances of models trained with component *a* or *b* separately is larger than that of phenomenon *T*, we can infer that the MUD likely uses disjunctive rules in inferring label *X*.

The challenge lies in building appropriate filters that can extract components of interest. Fortunately, recent advancements in computer science have enabled the development of tools that can be used for this purpose. For example, instance segmentation is one type of image segmentation that can be implemented with a deep learning model to not only identify the presence of a specific type of object (e.g., human) in image data, but detect, at pixel level, every instance of this type of object. This technology has greatly improved in recent years (Minaee et al. 2021) and is an essential tool in autonomous vehicles (Cordts et al. 2016), medical image analysis (Hesamian et al. 2019), and other critical applications. For image data, instance segmentation methods can be used as filters to extract interesting components, one at a time, from original phenomenon-level data. Instance segmentation models can be readily trained with a set of labeled data, requiring a one-time investment of time and effort by a researcher/practitioner. Many pre-trained models also are readily available with the ability to detect many common components, such as Facebook's DETR model, which has been trained on the challenging COCO 2017 object detection dataset (Carion et al. 2020).

The *ex ante* filtration module provides another path, in addition to the *ex post* experimentation module, for researchers and practitioners to assess which part of the original phenomenon-level data affects the outcome, and in which way. Conceptually, this is similar to the self-explicated method in preference measurement in marketing (Srinivasan and Park 1997) where a respondent is asked to state their preference, one feature at a time (i.e., a compositional approach), instead of being asked to state their preference for a complete product (i.e., a decompositional approach, such as conjoint analysis). This is the same logic behind incorporating the *ex ante* filtration module in the TMUD. The transparency of the MUD is achieved not by understanding how it processes complex phenomenon-level data, but by understanding how it reacts to already "purified" component-level data, one component at a time.

Comparison to Interpretability Research in Computer Science

The TMUD is related to current research in computer science on the interpretability of the deep learning model, which can be broadly divided into two types: feature attribution and model inspection (for an authoritative review of the deep learning field, see Raghu and Schmidt 2020). Feature attribution research (Simonyan et al. 2013, Selvaraju et al. 2017) focuses on determining which data features are most important in driving the model output (e.g., prediction), typically by using input masking to hide a single data point. For example, researchers studying tumor images typically block out certain pixels in an image to see how a reduced image affects the model's ability to classify a tumor, generating the so-called saliency map (Behrmann et al. 2018). Model inspection researchers (Olah et al. 2020, Bau et al. 2017) attempt to understand models by probing neurons in a neural network or varying input to evaluate which neurons are activated. Feature attribution involves investigating one input at a time using blunt force (local) masking, and model inspection involves investigating specific features of a neural network. In contrast, the proposed TMUD is a general framework that can be applied to any black box model (including neural networks) and can generate theoretical insights that are not possible with current interpretability approaches in computer science.

TMUD: Value and Relevance

As described in this section, with the addition of the two modules, the TMUD enables an investigator to uncover the inner workings of a black box machine model, thereby addressing the concerns of marketing researchers. Compared to a typical MAD, the TMUD offers substantially higher value in at least four aspects. First, the fit and predictive performance of a MUD is almost always better than a MAD (LeCun et al. 2015). This is expected, as a MAD only uses a subset of information in the model. This also has positive implications for theoretical investigations, as a model with better predictive capacity likely captures theoretical relationships more accurately as well. Second, inferences from a MAD, while straightforward (e.g., there is a positive relationship between the width of the mouth and perceived trustworthiness), only provide a partial perspective (e.g., width) on the role of a component (e.g., mouth). Unlike inferences from the TMUD, they do not reveal a component's overall (holistic) contribution to an outcome or the role of a holistic perceptual feature (e.g., sexual dimorphism). Third, the TMUD is much more flexible and efficient than a MAD. Prior to running a MAD, researchers must specify (a) which variable(s) they want to investigate, and (b) the functional form of the potential relationship(s) between the variable(s) and the dependent variable. On the other hand, the TMUD can investigate any factors, and uncover any functional forms and relationships, with no need to pre-specify variables (other than selecting the components in the *ex ante* filtration module) or functional forms. Possibly even more useful to researchers, the *ex post* experimentation module can be used after training the core MUD only once, and thus provide substantial efficiency. Fourth, MADs typically do not shed light on causation; rather, they provide evidence of correlation. The TMUD, on the other hand, allows a direct test of causation.

The TMUD is a timely and invaluable tool for researchers and practitioners, given the recent dramatic increase in the need to study unabridged data and understand the underlying mechanisms. Data suitable for the MUD have now become primary information sources for firms and are driving major marketing decisions. Many variations of unabridged data are readily collected and/or available to firms, especially text and visual data, and increasingly, audio data due to human interaction with voice-based tools like Siri. In addition, even for small firms, it is now easy to collect, store, and access vast amounts of unabridged data using readily available tools and cloud storage and computing facilities.

It should be noted that the TMUD was not feasible even a few years ago. Major elements in the two proposed modules have been developed quite recently, which makes it an even more timely development for researchers and practitioners. It is enabled by the emergence of powerful computing resources and breakthroughs in deep learning models. Computing power has improved dramatically, including advancements in hardware designed for deep learning (GPU, TPU), and these technologies also have become very affordable. Deep learning research has seen astronomic growth since the introduction of Alexnet in 2012 (Krizhevsky et al. 2012), with major breakthroughs almost every year. In addition, the deep learning discipline has encouraged (enforced in many cases) the practice of sharing code and even data and/or pretrained models with researchers, greatly accelerating diffusion. Major corporations such as Nvidia, Google, Facebook, and Microsoft also have invested tremendous resources into research, and often make their research results public, including generator models (StyleGAN/StyleGAN2) and instance segmentation models that are critical to the TMUD.

## An Empirical Investigation of Faces in Marketing Contexts

In this section, we present an empirical application designed to demonstrate how the TMUD can be used to better understand the inner workings of a black box model used to predict marketing outcomes. First, we describe the important role of face perception in several key marketing contexts. Then, we describe the data, the black box MUD, and our analysis before presenting and discussing our results.

Built on these data and the trained MUD, the next two sections shed insights into these models using the prescribed analysis under TMUD. Specifically, the next section explores the holistic roles of various facial components using the *ex-ante* filtration module of TMUD. The section after it investigates how sexual dimorphism contributed to such perceptions using the *ex post* experimentation.

Face Perception in Marketing

It is well documented that consumers react strongly to faces in commercial contexts, such as the faces appearing in advertisements and on websites, and the faces of service providers (e.g., real estate agents, lawyers) (Kahle and Homer 1985, Rule and Ambady 2008, Gorn et al. 2008, Xiao and Ding 2014). While such effects are well documented, they are based on limited MAD outputs: either geometric and chromatic facial attributes or specific facial features (Vernon et al. 2014, Todorov et al. 2008), or principal component-based features such as eigenfaces (Xiao and Ding 2014). These models are clean, and the geometric/chromatic attribute approach provides easy-to-interpret results.

However, MADs have four major limitations, as discussed in the previous section. First, the fit and predictive performance of these models are good, but not great. Second, inferences regarding how such perceptions are made, while straightforward, only provide partial information about the role of each component (e.g., there is a positive relationship between the width of the mouth and perceptions of trustworthiness). They do not reveal the bigger (holistic) picture regarding how humans form perceptions (i.e., how important is the mouth to overall perceptions of trustworthiness). Third, they are limited by specifications of engineered features and functional forms of potential relationships (how each attribute affects the perception), which could easily be mis-specified or simplified. Fourth, these MADs do not shed light on causation; they only provide information about correlation.

We present an empirical application to explicitly showcase how the TMUD overcomes these limitations in four major marketing contexts with a total of six perceptual dimensions (or labels, as typically called in deep learning literature) where face perceptions are critical: advertisements (perceptions of trustworthiness, competence, and approachability) (Xiao & Ding 2014), brand (perceptions of whether a face represents a brand's typical customer) (Tkachenko and Jedidi 2020, Liu et al. 2020), category (perceptions of whether a face represents a typical customer of the product category) (Tkachenko and Jedidi 2020), and customer persona (perceptions of whether a face represents the persona of a brand's customer segment) (Pruitt and Adlin 2010, Holzwarth et al. 2006, Lester et al. 1997).

While the first three contexts are intuitive and familiar to most researchers and practitioners, customer persona is complex and requires further elaboration. Customer persona is the hypothetical archetype of real users (Pruitt and Adlin 2010) with specific characteristics (e.g., gender, age, occupation, consumption habits, hobbies, etc.). Firms develop unique and distinct customer personas by identifying the preferences and needs of their target customers and using appropriate channels to reach and retain them (Forbes Agency Council 2018). In addition to a textual description, a customer persona is usually accompanied by a face image to make it more vivid and concrete (Lester et al. 1997). Therefore, designing an appropriate face image for a customer persona not only helps focus internal communication about target customers to facilitate the development of creative marketing strategies (So and Joo 2017), but also helps companies deliver effective messages to the market (Holzwarth et al. 2006).

The purpose of TMUD application here is to first develop an accurate (black box) MUD that can predict which face will be most appropriate in each of these contexts (this section). This is followed by additional analysis to understand how different facial components (next section) and sexual dimorphism (the section after next) influence whether a face will be considered appropriate in these contexts, thus shedding lights to how these black box MUD works.

Data

Following standard deep learning research on images, we curated a large dataset of faces that are appropriate for the marketing contexts under study, and then recruited research assistants (RAs) to apply each of the six labels to each face. Due to Covid-19 restrictions, we recruited the human evaluators in Shanghai (where one of the authors was located) where it was possible to engage in in-person training and monitoring in 2020.

We obtained face images from public LinkedIn pages and screened them, retaining only high resolution images showing a front view of the face, with no face-blocking items such as sunglasses. We also excluded images that had clearly been enhanced (i.e., Photoshopped). We decided to only collect faces that of college-age individuals (current students or recent graduates) because one of the marketing contexts we wanted to study was the Starbucks persona for a college student customer segment. Because the RAs were based in Shanghai, we limited images to Asian faces. It is well documented that humans are not very good at making inferences about people from different cultures based on their appearance (Wolf 2013). After screening, we retained 3,383 images (female: 1,931; male: 1,452). To avoid distractions of hairstyle and jewelry, all images were cropped to exclude hair and ears before they were labeled by the research assistants.

We defined the six labels for the four marketing contexts as follows. For the three perceptual labels in the advertisement context (perceptions of trustworthiness, competence, and approachability) we used definitions found in the literature (Vernon et al. 2014, Xiao and Ding 2014). For the brand context, we defined the label as "a person who looks like a regular customer of Starbucks." For the product category context, we selected lab meat (i.e., meat products produced using *in vitro* cell cultures of animal cells) to understand consumer perceptions of a disruptive new product category. Although lab meat helps address the challenges of global food security and climate change, many people are hesitant to try it. Thus, we defined the label as "a person who looks like they are likely try lab meat." Finally, for the customer persona context, we followed the principles of customer persona design in marketing practice (Babich 2017) and developed a customer persona of college students who patronize Starbucks as follows: *A junior in college, liberal arts major, originally from southern China; extrovert, likes to try new things, likes to have control, very confident and has own opinion, blunt, quick learner; often goes to Starbucks to meet up with friends for casual gatherings, and typically goes there after class, with 3–5 friends; very familiar with the various drinks offered by Starbucks*. We defined the corresponding label as "a person who looks like someone in this customer segment."

We recruited and trained 21 RAs (9 males, 12 females) to independently apply the six labels associated with the four marketing contexts to each face using a scale from 1 (absolutely does not look like the label) to 5 (absolutely looks like the label). They were instructed to rely on gut instinct and assured that there were no right or wrong answers. In addition, they were asked to identify the gender of each face (female: 0; male: 1). We used the gender labels in the *ex post* experimentation module to investigate the effect of sexual dimorphism on face perceptions. Following standard practice in the literature (Vernon et al. 2014), we used the probability of being perceived as male/female as a proxy for sexual dimorphism.

The RAs performed all face labeling work in Qualtrics after receiving personal training from one of the authors. The authors randomly grouped the 3,383 faces into eight sets of 420–443 faces. RAs were asked to rate all faces in a set for one label before moving on to the next label to minimize possible artificial correlation among the labels. The order of the labels was randomized, as was the order of the photos. To assess the reliability of the ratings, the authors randomly duplicated 20 face images in each questionnaire. The correlation coefficients of the 20 repeatedly scored face images indicate the test-retest reliability of an RA's rating. This practice is consistent with studies in which 10 or more raters assign labels (Todorov et al. 2013, Oh et al. 2019). If one of the seven labels rated by a RA for a given questionnaire had a negative or zero test-retest reliability, all data from the RA's questionnaire were excluded (Oh et al. 2019). Based on the above criteria, the RAs provided 295,771 valid ratings, with an average of 12.49 ratings per dimension (label) per face image, with a standard deviation of 1.44. We calculated the average rating for each face on a specific label, assigning a code of 1 to scores greater than 3, and 0 otherwise. For gender, we assigned a code of 1 (male) to an average score greater than 0.5, and 0 (female) otherwise. These codes constituted the final labels.

All face images had a length and width of 224 pixels, with each pixel represented by three integers ranging from 0 to 255 for the three channels of RGB (red, green, and blue). This matrix of (224, 224, 3) was scaled to (0, 1) before being used to train the models.

Model and Analysis

The models we used to generate insights into how consumers perceive human faces along the six perceptual dimensions and gender are the so-called classifiers in deep learning literature. Following state-of-the-art practice, we used convolutional neural network (CNN) models as the classifiers (i.e., the MUD used in this empirical application). CNN is the go-to technology for processing and analyzing image data, and has excellent performance in image classification, retrieval, target segmentation, face recognition, medical image analysis, and so on.

Adopting the transfer learning approach (Bengio 2012, Zhuang et al. 2020) in deep learning research, we used a pre-trained CNN model called VGGFace (Parkhi et al. 2015). We unfroze the last block so that weights could be updated during training, and replaced the last layer's activation function, Softmax, with the Sigmoid activation function. We then trained the model against our labels, one at a time.

Prior to model training, data in each of the seven datasets were randomly divided into a training set (70%), a validation set (15%), or a test set (15%). The images within each dataset were then balanced to ensure an equal number of face images in both classes of a label. For example, if the number of face images coded as trustworthy had been greater than the number of images coded as not trustworthy, we would have kept all images labeled not trustworthy, and retained an equal number of images labeled trustworthy, eliminating the rest, beginning with images whose average ratings were furthest from their labels (0 or 1).

Model training was performed on the Google Colab with GPU. Data were augmented (Shorten and Khoshgoftaar 2019) by flipping images horizontally, and adjusting the scale, zoom, brightness, and shear angle. We also used early stopping to avoid overfitting. If the loss function of the model on the validation set did not improve for five consecutive epochs (at least 0.1%), the training was stopped and the best model parameters were saved. On average, it took 20 minutes of server time to train each model.

Results of MUD

The predicted performances of the seven trained MUDs are included in Table 3 (refer to the row labeled Phenomenon-Level). Model performance is indicated by the predictive accuracy of each label (i.e., the number of images correctly predicted divided by the total number of images in the test set).

Insert Table 3 here

For the labels that have been studied in the literature, our results are consistent with those for similar deep learning models. The trained MUD is very accurate in predicting the gender of a face (all hair and jewelry were blocked from the images), at 99.5% accuracy (similar to VGG-16, Lapuschkin el al. 2017). Results for the three perceptual dimensions often studied in the advertisement context are also consistent with the literature. Messer and Fauser (2019) inferred warmth and competence from face images with an accuracy of about 90% and 80%, respectively. Our model predicts approachability, trustworthiness, and competence with 89.7%, 80.8%, and 83.4% accuracy, respectively.

These models have not been used to predict face perception in the remaining three marketing contexts in prior studies. Our results indicate that humans are able to identify faces that are most suitable for each of these contexts, and that our trained model captures this phenomenon. The model can predict whether a person looks like a regular Starbucks customer with 84.8% accuracy. More interestingly, the model can predict early adopters (innovators in the product diffusion literature) for a new product category (lab meat) with similar accuracy, at 83.2%. Finally, the model is equally capable of predicting whether a face fits with a brand persona for a specific customer segment (i.e., a college student who patronizes Starbucks) with 82.4% accuracy.

These results provide further evidence that the MUD is extremely capable of making predictions in a diverse set of marketing contexts, although they do not tell us how it makes such predictions. In the next two sections, we use the *ex ante* filtration module and *ex post* experimentation module of the TMUD to demonstrate how researchers can open these black box models and obtain theoretically meaningful and actionable insights.

## Role of Facial Components on Face Perception

Built on the analysis described in the last section, this section explores the holistic roles of key facial components in face perception using the *ex ante* filtration module of the TMUD. We first describe the *ex ante* filter used, and after a brief description of the model and our analysis, we examine the results and discuss how they shed light on how humans make the six facial perception inferences in the four marketing contexts using each of the core facial components.

**The *Ex Ante* Filtration Module**

To ensure the *ex ante* filtration process is effective, filters must be identified to extract the important components that are of theoretical interest. In this empirical application, we were interested in major facial components: mouth, nose, eyes, eyebrows, and face contours. Several published models serve this purpose well (Liu et al. 2017a, Smith et al. 2013). We decided to use a recent model, BiSeNet (Yu et al. 2018) which has received a very positive reaction from the deep learning discipline (435 Google citations within 3 years of its publication). We applied the BiSeNet model, which was pre-trained on the CelebAMask-HQ dataset (Karras et al. 2017), to extract eyes, mouth, nose, eyebrows, and face contours for each image using Google Colab (see Figure 5 for an example).

Insert Figure 5 here

We created 6 x 7 = 42 additional datasets, where 6 refers to the five facial components plus a composite face constructed using the extracted five facial components only (see Figure 5); and 7 refers to the seven labels: gender (sexual dimorphism), approachability, trustworthiness, competence, Starbucks persona, perceptions of Starbucks customers, and perceptions of lab meat customers. The value of a label for the original face carries over to its facial components or the composite face.

**Model and Analysis**

These 42 datasets were used to train the VGG-Face model described in the previous section, following the same process.

**Effect of Facial Components**

The predictive performances of these 42 models are reported in Table 3, in rows labeled "Component-level." This analysis yields two major insights. First, it reveals the role of holistic facial components in facial perception. To the best of our knowledge, we are the first to identify the unique contribution of a holistic facial feature (e.g., a mouth) to facial perception. This adds substantial value to literatures across many disciplines ranging from computer science to psychology in which scholars have only studied the contribution of attributes of facial components (e.g., width of a mouth) in isolation. In addition, our results show that different facial components play very different roles across the seven perceptual dimensions, ranging from high to moderate to low to no effect. Second our results reveal the non-compensatory (disjunctive) nature of facial perception. We show that facial perceptions based on only one or a subset of facial components typically are the same as those based on the original face. It can also be inferred that some components' contributions to a specific facial perception are redundant. Thus, it is possible to implement models using only a subset of facial components in applications where either the entire face image is not available (e.g., someone wearing a mask) or a certain part of the face needs to be blocked to remove bias (e.g., race, gender). We elaborate on these two insights in detail.

The contributions of facial components to face perceptions are significant and heterogeneous, both across perceptual dimensions and across features. Models trained using only the mouth feature predict all face perceptions well, ranging from 62.6% (perception of competence) to 82.6% accuracy (perception of approachability). This points to the critical role the mouth plays in face perception in general. Similarly, face contours appear to be very important in face perception, as it alone can predict competence with 67.3% accuracy and gender with 93.8% accuracy. The role of the nose is also relatively consistent across all perceptions, with accuracies around 70%, just slightly less than the mouth and face contours, except it appears to contribute no information to the perception of competence. The role of the eyebrows is more heterogeneous, ranging from quite important (gender: 81.9%), to somewhat important (perception of approachability: 65.0%; trustworthiness: 58.5%; frequent customer at Starbucks: 62.0%), to not informative at all (perception of competence, Starbucks persona, and lab meat customer). At the other extreme, eyes appear to play no role in any of the face perceptions. This result, however, must be interpreted with a caveat. The eyes we studied only include areas between the inner and outer eyelids, and did not include other information such as eyelids, eyelashes, eye bags, etc., which are typically part of eye perception. Because we know that eyes play an important role in face perception from our personal experience, this indicates that the role of the eyes is not driven by the sclera (the white of eyes), iris, or pupil. It should be noted, however, that all face images used in this study were Asian, and there was no variation in the color of the iris (all were brown).

If we examine the contributions of different facial components to a given face perception, it becomes clear that humans use non-compensatory (disjunctive) rules to make such inferences. More than one facial feature informed each perceptual dimension (label) quite well. More tellingly, the sum of correct predictions from all facial components for a given face perception is much greater than 100%. In some cases, a single facial feature generates predictions that are almost as good as the original face image.

Four facial components (mouth, nose, face contours, and eyebrows) independently contribute to the correct prediction of gender, with accuracy ranging from 69.7% (mouth) to 93.8% (face contour). In terms of approachability, the mouth and face contours are the main basis for judging the impression of approachability, with predictive accuracies of 82.6% and 75.3%, respectively. The predictive accuracies of the eyebrows and nose were slightly lower, at around 65%. The mouth, face contours, and nose are the main basis for perceptions of trustworthiness, with about 70% predictive accuracy. Although eyebrows can predict trustworthiness, their predictive accuracy is only slightly higher than the probability of random prediction, at 58.5%. Only two facial components can help predict perceptions of competence: the mouth (62.6%) and face contours (67.3%). This is consistent with the literature, as Messer and Fausser (2019) showed that the area under the eyes and on both sides of the nose are important areas for the formation of competence impressions. Three facial components (mouth, nose, and face contours) are informative in perceiving whether a person fits the given Starbucks persona for the college student customer segment, with a predictive accuracy ranging from 68.6% (nose) to 76.1% (mouth). As for whether a face looks like a regular Starbucks customer, three features (mouth, nose, and face contours) provide substantial information (with accuracy between 70.7% and 73.1%), with eyebrows as a distant fourth (62% accuracy). Similar to the Starbucks persona, three features can convey whether a person is perceived as a likely customer for the lab meat product category. Among them, the mouth is the most informative, with a predictive accuracy of 77%. The nose and face contours have a predictive accuracy of about 67%.

Finally, the composite face, constructed using the five facial components, performs better than any individual feature and worse than the original face image, as expected. What is surprising, however, is that the composite face performs only slightly worse than original face: 1–4% worse on six dimensions/labels, and 8% worse for competence. Therefore, it would be very advantageous to use composite faces to predict consumer perceptions in practical applications, as it would yield better results than models using a single facial feature (such as the mouth) while preserving the ability to help reduce cognitive biases by eliminating certain information from images.

## Role of Sexual Dimorphism on Face Perception

The TMUD's *ex post* experimentation module provides complete freedom to investigate how a trained MUD model makes inferences about a particular dependent variable (label). In this section, we explore the role of sexual dimorphism in people's perceptions of faces on the six dimensions using the *ex post* experimentation module and demonstrate how we can uncover the role of an explanatory variable after the model has been trained without pre-specifying the functional form of how this might affect the dependent variable.

### Literature on Sexual Dimorphism

We chose sexual dimorphism (Gangestad and Scheyd 2005) as a variable to demonstrate the TMUD's ability to uncover a causal relationship that drives the inner workings of a black box model. A key construct studied in the face perception literature, sexual dimorphism is an important facial feature which can affect other face perceptions. There is a large body of research on the relationship between sexual dimorphism and approachability, trustworthiness, or competence perceptions, but no prior research on its relationship with perceptions of a brand, category, or persona, which are idiosyncratic.

Based on the literature, the relationships between sexual dimorphism and approachability or trustworthiness are rather straightforward. From an evolutionary perspective, faces can help us detect potential threats in human interactions. It has been shown that facial masculinity is positively correlated with perceived dominance and aggression (DeBruine et al. 2006, Oosterhof and Todorov 2008). As a result, people tend to perceive faces with strong masculine features as less approachable and less trustworthy (Hess et al. 2004 , Sutherland et al. 2013).

The literature on the relationship between sexual dimorphism and competence is much more complex, and is strongly influenced by both evolutionary considerations and the culture in which such inferences are made. Existing literature on this topic is almost exclusively based on experiments/observations of individuals in Western cultures, and findings show that masculinity is a main driver of impressions of competence (Oh et al. 2019, Sutherland et al. 2013, Broverman et al. 1972, Walker and Wänke 2017). This is consistent with evolution theory, as masculine characteristics are important manifestations of physical strength (Fink et al. 2007), and thus indicate competence (and significant advantages) in environments where physical strength is critical.

However, well-documented cultural differences in face perception (Sofer et al. 2017, Zhan et al. 2021) might counteract those driven by evolution. Cultural differences in perceptions of competence are particularly prominent between East Asian cultures (for example, the greater China region, Japan, and Korea) and Western cultures (for example, North America and European countries). In contrast with Western cultures, faces with masculine features in East Asian cultures are not necessarily perceived as more competent. Whereas people in both cultures associate those who can undertake demanding physical work with more obvious masculine facial features, individuals in East Asian cultures traditionally value cerebral work and look down upon people who do physical work. This is exemplified by a poem written by Chinese Emperor Zhenzong (AD 968–1022) of the Song Dynasty that evolved into a proverb and influenced young people for more than a thousand years: "In books, you can find houses made of gold and people beautiful like jade." In other words, wealth and a beautiful spouse will come to anyone who masters knowledge (books). In East Asian cultures, competence (even eminence) is reflected in one's intellect (Shen 2018) rather than physical prowess. People in these cultures even stereotype those with distinctly masculine traits as "strong-bodied but simple-minded."

This is reinforced by the civil service examination system, in place from the mid-Tang dynasty to the late Qing Dynasty (almost 2,000 years) in China, which was the only way for people to climb the societal hierarchy. This singular path for upward mobility has cemented a cultural definition of competence as equivalent to intellect; physical strength is not relevant. Even today, high schoolers in China, Japan, and Korea take grueling national college entrance examinations that determine which colleges they attend, which in turn largely determines their future careers. More recently, under the influence of Japanese and Korean pop culture, which celebrates non-muscular (even feminine-looking) males, many young people in East Asia now prefer males with more feminine features (mouth, eyes, nose, facial contours, eyebrows, hairstyle, skin color, etc.) (Zheng 2015, Shen 2018). Similarly, in East Asia, females are perceived to be more competent if they look more feminine due to traditional role expectations which do not involve physically demanding work. In fact, looking more masculine is often perceived as a negative for females in society.

Because we used Asian faces exclusively and all of the RAs were current students in East Asia, we anticipated that these evolutionary and cultural forces would counteract each other, and the results for the relationship between masculinity and perceptions of competence would depend on the extent to which evaluators were influenced by culture and the faces being evaluated (some faces could be considered more from the cultural norm angle, while others could be considered more from the evolutionary angle). We thus anticipated heterogeneity across both Asian observers and Asian faces. In other words, some people may perceive feminine faces as more competent, whereas others may perceive them as less competent. Also, some faces may be perceived as more competent when they become more feminine, whereas others may be perceived as less competent.

**Data Generation for Causal Relationship Testing**

To provide a causal explanation for the role of sexual dimorphism, we needed to create a large set of stimuli for which we could control all aspects except the degree of sexual dimorphism. Data generation involved three steps. First, we encoded the original images in the StyleGAN latent space. Second, we identified the latent space direction along which we could vary the factor (sexual dimorphism) under study. Finally, we modified the original images by moving their latent space vectors along the latent space direction identified in step two and generated many variations of each original image that differ gradually only in sexual dimorphism.

The encoding is completed as follows. First, we encoded the 3,383 faces in our dataset in StyleGAN's latent space (Karras et al. 2019) to obtain the corresponding latent space vectors for each face. We then regenerated faces based on these latent space vectors and compared them to the original face images, retaining only the latent space vectors that could regenerate faces indistinguishable from the corresponding original faces. This left us with a total of 2,502 face images; most eliminated images were of individuals who were wearing glasses. Since we planned to modify sexual dimorphism using the original faces, we decided to use only those that were unquestionably male (probability of male > 0.99) or female (probably of male < 0.01), leaving us with a final set of 2,314 faces (female: 1,619; male: 695).

We identified the sexual dimorphism direction in the StyleGAN latent space following procedures in the literature (Shen et al. 2020, Karras et al. 2019). The gender prediction model was trained using the support vector machine (SVM) method, with an accuracy of 99.3%. The trained SVM returned a latent space vector representing the sexual dimorphism direction which is orthogonal to the male/female hyperplane.

Finally, we edited each original face along this direction using the face's latent space vector to create images that differed from the original faces only in gender perception (i.e., female faces became more masculine, male faces became more feminine). We evaluated these edited faces to ensure they looked realistic and were indistinguishable from real face images. We generated 40,518 modified face images for our study, but we could have generated many more if necessary.

Although the latent space vector can help us adjust the gender of faces, scholars are still unclear about the quantitative relationship between the change in a latent space vector (of face images) and the corresponding change in label (gender). As a result, we fed these edited faces into the already trained VGG-Face on gender to estimate the probability of being male for each edited face to measure the degree of sexual dimorphism. Figure 6 shows two examples of face editing, one male and one female. Only very subtle differences can be detected by the naked eye, and yet gender classifiers perceived substantial differences in sexual dimorphism, supporting the validity of the latent space direction of gender. Note that for privacy reasons, these two faces were not used in the study.

Insert Figure 6 here

To facilitate our investigation of the effect of sexual dimorphism on face perception, we reduced this set of edited faces as follows. Since we wanted to study the effects of sexual dimorphism when the gender identity of the original face does not change, we only retained edited faces with the same predicted gender as the original face (i.e., predicted male probability: 50–100% if the original image was labeled male; <50% if the original image was labeled female). We also balanced the datasets so that each 5% interval (predicted probability of being male) included a comparable number of edited faces by reducing the number of images in the overrepresented intervals. The final dataset of edited faces included 17,084 faces: 5,176 edited faces based on the original 695 male faces, and 11,908 edited faces based on the original 1,619 female faces.

**Analysis**

Once we obtained the edited faces that only differed on the sexual dimorphism dimension, the next stage in *ex post* experimentation involved three steps. First, the edited faces were fed into the trained MUD to estimate how each face was perceived on the sexual dimorphism dimension (label). Second, each edited face (image) was compared to the original image to obtain a difference measurement indicating how a change in the perceived sexual dimorphism of a given edited face affects a change in the label. Finally, we used statistical tools to analyze these difference measurements in order to identify the specific causal relationship. We discuss each step in more detail below.

We fed the 17,084 edited faces one at a time into the six MUDs that we had trained to capture six important face perceptions in four marketing contexts. We used these trained MUDs to predict the probability that the person depicted in the edited face image is approachable, trustworthy, competent, a Starbucks persona, a typical Starbucks customer, or likely to try lab meat.

We used these predicted probabilities, together with the predicted probabilities of being male/female (sexual dimorphism), to create difference measures for each original face image to yield a clean assessment of the effect. For each edited face on a specific perception (label), we calculated the difference in the probability of being perceived as male/female between the edited face and the original face and used this as the *x*-value (independent variable). We then calculated the difference in the probability of being assigned the label between the edited face and the original face and used this as the *y*-value (dependent variable). Whereas the *y*-value depends on the specific perception being investigated, the *x*-value does not.

For ease of interpretation, we analyzed male faces in terms of feminization. For male faces, the *x*-value is the extent of feminization (i.e., the difference in the probability of being female between the edited image and the original image). Likewise, we analyzed female faces in terms of masculinization. For female faces, the *x*-value is the degree extent of masculinization (i.e., the difference in the probability of being male between the edited image and the original image).

Finally, we applied statistical analysis to these difference measures to identify whether a causal relationship exists between sexual dimorphism and a specific perceptual label, and if so, what the specific relationship is. We tested both linear (Equation 1) and quadratic (Equation 2) relationships, but a wide variety of possible functional forms could be investigated, as the TMUD can generate an infinite amount of data to capture even unusual relationships. We used the following equations:

$$y_i = a_{0i} + a_{1i}x \qquad (1)$$
$$y_i = b_{0i} + b_{1i}x + b_{2i}x^2 \qquad (2)$$

where *i* refers to perception (label), and *a* and *b* are corresponding coefficients.

In the next two subsections, we first present the aggregate level results, followed by individual level results aimed at understanding the heterogeneous nature of the causal relationship. The individual level results are organized by individuals who evaluated the faces (i.e., the RAs) and by faces that were evaluated.

**Aggregate Level Effect**

Table 4 shows the aggregate level effects. We report the value and significance of the coefficients for the first- and second-degree polynomials as well as the adjusted R-squared values. We discuss the effect of sexual dimorphism for male faces before discussing this effect for female faces.

Insert Table 4 here

As discussed in the literature review in this section, existing literature indicates a positive correlation between feminization and perceived trustworthiness and approachability for male faces. Results of our controlled experiment show strong support for these correlation-based hypotheses by revealing that the causal relationships are quadratic. The literature on competence is more complex due to the counteracting effects of evolution and culture, as discussed earlier, and we hypothesized this to be highly dependent on the evaluators (the RAs in our case) and the faces being evaluated. This turned out to be the case, with no significant polynomial coefficients and an extremely low adjusted R-square value. Researchers have not yet explored the effects of sexual dimorphism on perceptions that an individual is a specific brand's customer, a specific brand persona for a segment, or a likely customer for a new product category. Furthermore, these effects are likely to be highly context-dependent. As shown in Table 4, quadratic relationships exist between sexual dimorphism and perceptions of a face reflecting the Starbucks persona and a typical Starbucks customer. On the other hand, the effect of sexual dimorphism on perceptions of a face reflecting a lab meat category customer is linear, with an insignificant quadratic term.

The effects of sexual dimorphism for female photos are largely consistent with those for male photos, with two exceptions. Since female faces were analyzed in terms of masculinization, a negative coefficient in the linear term for female faces is equivalent to a positive sign for male faces, which were analyzed with regard to feminization. First, female faces that are made more masculine are perceived to be less competent, indicating that cultural factors play a stronger role than evolutionary factors in determining perceptions of female faces for the RAs who assigned the labels. However, for male photos, sexual dimorphism has no effect on the perception of competence. Second, the effect on perceptions of lab meat customers is best captured in a quadratic functional relationship, instead of the linear relationship for male faces.

**Heterogeneity in Effect**

To provide a better understanding of the role of sexual dimorphism, we investigated the individual heterogeneity across evaluators (RAs) and across face images. It is quite possible that individuals form unique perceptions of a face on a particular dimension; for example, a face that looks trustworthy to one person may not look trustworthy to another person. This difference is mostly due to differences in life experiences, cultural background, and genetic traits (Sofer et al. 2017, Zhan et al. 2021). Therefore, sexual dimorphism may have had different effects on different evaluators' perceptions of faces for the six dimensions (labels) studied. Although the number of RAs in the study was not sufficient to draw general conclusions, we explored potential differences by analyzing the results of two RAs. We trained the six MUDs using the labels assigned by each RA (rather than the average of the RA pool), and re-ran the analysis described in this section. The results are shown in Table 5.

Insert Table 5 here

The magnitude of effects between RA1 and RA2 and the aggregate results show substantial differences. More interestingly, qualitative differences are indicated by the difference in the signs of the polynomial coefficients for the same face perception dimension (label). The main difference is the role of sexual dimorphism in the perception of competence, which is not surprising, given the counteracting forces of evolution and culture. Although the feminization of male faces has no significant effect on the perception of competence at the aggregate level, RA1's perception of competence decreased ($p<0.001$), whereas RA2's perception of competence increased ($p<0.05$) as feminization increased. Similarly, as the masculinization of female faces increased, RA1's perception of competence increased ($p<0.001$), whereas RA2's perception of competence decreased ($p<0.001$). These results seem to indicate that RA1 was driven more by evolutionary factors whereas RA2 was driven more by (East Asian) cultural factors when inferring competence based on sexual dimorphism.

It is conceivable that heterogeneity exists across faces. Some male faces appear to be more competent if they look more feminine, whereas other male faces appear to be less competent if they look more feminine due to the unique features of the origin faces. We investigated this possibility by running the analysis at individual (original face) level, using the original face images and their edited copies, which varied only on sexual dimorphism. To ensure the model could be estimated with reasonable accuracy, we only analyzed original faces with eight or more edited copies in our dataset, leaving us with 1,050 original faces (male: 333; female: 717) and 9,054 total images (male: 2,922; female: 6,132). The signs of the first-order derivatives of the quadratic models are almost the same as those of the linear models; thus for simplicity, only the results of the linear models are reported here. Table 6 presents the values of the coefficients ($a_1$) for males and females in the linear functions, divided into three categories based on their signs: negative, positive, and insignificant ($p>0.05$).

Insert Table 6 here

The first observation from this analysis is that the effect of sexual dimorphism is quite robust across different faces for four face perceptions: approachability, trustworthiness, Starbucks persona, and Starbucks customer. In all cases, perceptions of 89.2–98.2% of individual faces are consistent with the aggregate level results, and with extremely low opposite effects (0.6–3.9% of individual faces). The second observation tells a very different story, not surprisingly, about perceptions of competence. Sexual dimorphism appears to lead to substantially different inferences about competence based on which face is being evaluated. For male faces, the effect of being more feminine is negative for 16.8%, positive for 59.8%, and insignificant for 23.4%. For female faces, there is less heterogeneity, but significant separation still exists between the effects of masculinity, which is negative for 75.9%, positive for 9.1%, and insignificant for 15.1%. Finally, the effect of sexual dimorphism on the perception of a lab meat customer is slightly more heterogeneous than its effect on perceptions of approachability, trustworthiness, Starbucks persona, and Starbucks customer, especially for male faces, but is much more homogeneous compared to that of competence.

## General Discussion

MUDs (such as deep learning models) have enjoyed tremendous growth in business applications in recent years for two reasons. First, we are now in the era of rich unstructured data (e.g., images, videos, texts) and firms are increasingly relying on insights from such data to guide their managerial decisions in order to successfully compete in the market. Second, computer scientists (in both academia and the tech industry) have made major strides in methods (software) and computational power (hardware) which have made the analysis of unstructured data not only feasible, but also accurate and efficient. Nevertheless, marketing scholars have largely stayed away from such MUDs because they operate as black box models; they work well, but we do not know why they work. The aim of this paper was to help resolve this concern and substantially increase the adoption of MUDs in the marketing discipline.

The proposed transparent model of unabridged data (TMUD) synthesizes scholarship from marketing, computer science, and electrical engineering, and incorporates an *ex ante* filtration module and an ex post experimentation module to enable researchers to uncover the internal mechanisms of these black box models to gain theoretical insights. We have illustrated an application of the TMUD using six dimensions (labels) of face perception in four important marketing contexts. We have shown how the *ex ante* filtration module can be used to identify the roles of five facial components (mouth, nose, eyes, eyebrows, and face contours) in determining overall face perceptions in each of these contexts. Interestingly, we also found that humans likely adopt disjunctive decision rules in face perception, and that perceptions based on only one facial component are sometimes almost as accurate as perceptions based on the entire face. Using the *ex post* experimentation module, we have shown that it is possible to identify the role of a specific holistic construct (i.e., sexual dimorphism) on these face perceptions. One main finding of this analysis is that the effects of sexual dimorphism on perceptions of competence are complex, and depend on the faces and the evaluators, likely due to counteracting evolutionary and cultural forces.

We expect that the TMUD will help advance marketing research and practice in three ways. First, it will open the door to scholars and practitioners who are interested in investigating theoretical insights using unstructured data. Second, it will encourage the adoption of MUDs in general, because it eliminates researchers' and practitioners' concerns that they will not be able to explain MUDs to their audiences (colleagues, clients, supervisors) if they use them, even if they are mostly interested in outputs (e.g., predictions) rather than why the models work. Third, it can be used to help screen and/or design the most desirable unstructured data points (e.g., select a specific face from a set of models, design an artificial CGI character, or design a specific exterior shape of a car) in a given marketing context (e.g., a face used in a print advertisement for a brand), once we understand which component and specific characteristics of that component and which holistic feature (e.g., sexual dimorphism) boosts overall perceptions of the phenomenon.

Our findings reveal several fruitful and important directions for future research. First, due to space limitations, we did not combine *ex ante* and *ex post* modules in our analysis. This would be useful to scholars who are interested in specific components to study, for example, the sexual dimorphism of the mouth on perceptions of competence. Second, because research on latent space direction is still a very active domain in computer science, it will be important to extend methods that enable users to obtain a clean direction in latent space for a holistic theoretical construct of interest. Third, applications in additional empirical contexts would support practical utilization of the TMUD. In our empirical study, we used faces as the unstructured data and studied face perceptions in four major marketing contexts; however, the TMUD can be used to study any type of data, including digital (e.g., print ad design) or physical (e.g., human, car exterior, clothing, interior design, product package) objects, as long as they can be photographed accurately, as well as non-visual data such as audio and text. Applying the TMUD in these diverse domains will improve best practices and convince more researchers and practitioners to utilize it. Finally, we used Chinese coders in Shanghai due to Covid-19 restrictions on human interactions in 2020. It would be very interesting to re-run the empirical study with people from other cultures, especially given our finding that perceptions of competence are heavily influenced by culture.

To the best of our knowledge, the TMUD is the first systematic framework that can guide researchers and practitioners to uncover the inner workings of a black box MUD. We hope its application and acceptance will substantially enrich marketing theories while improving practices in the field.

Table 1. Data Types with Examples

| Data type | Unabridged data (P: phenomenon level; C: component level) | Abridged data (P: phenomenon level; C: component level) |
|---|---|---|
| Visual: Face | P: A picture of the face<br>C: All pixels of mouth | P: Geometric measurements of all facial features, and color<br>C: Geometric measurements of mouth |
| Visual: Car exterior | P: Pictures of the car that capture all angles of the exterior<br>C: All pixels of the taillights | P: Geometric measures of all major exterior components (e.g., windows, doors, hood) and color<br>C: Geometric measurements of doors |
| Visual: Print ad | P: An image of the print ad<br>C: All pixels of the dog used in the ad | P: The presence of people (and their demographics), objects (such as a sofa), color theme, implied story (category variables), etc.<br>C: Number of people present |
| Audio: Song | P: Audio recording of the song<br>C: The recording of the trombone used in the song | P: Coding of the song on a set of variables such as genre, singer, tempo, rhythm, instruments, etc.<br>C: Profile of the singer (age, gender, voice register) |
| Smell and taste: Wine | P: Actual wine smelled and tasted (full chemical profile of the wine)<br>C: Chemical profile within wine that generates the aroma | P: Attributes of wine such as different types of fruit, degree of sweetness, etc.<br>C: Degree of sweetness on a scale of 1–100 |

Table 2. Model Types with Examples

| Type | MAD (Model of abridged data) | MUD (Model of unabridged data) |
|---|---|---|
| Biological (human, mind) | An adult learning to detect lies after being taught to observe certain eye movement patterns | Babies learning to read facial expressions by themselves |
| Machine | A model learning to predict how soon a dog will be adopted based on breed, weight, height, hair and eye color, etc. | A deep learning model learning to tell whether there is a dog in an image by examining numerous images with or without dogs |

Table 3. Predictive Performance of Trained MUD: Full Face and by Component

|  | Sexual dimorphism | Approachability | Trustworthiness | Competence | Starbucks persona | Starbucks customers | Lab meat customers |
|---|---|---|---|---|---|---|---|
| Number of images in the test set | 436 | 340 | 314 | 212 | 472 | 376 | 392 |
| Phenomenon level | | | | | | | |
|   Full face | 99.5% | 89.7% | 80.8% | 83.4% | 82.4% | 84.8% | 83.2% |
| Component level | | | | | | | |
|   Mouth | 69.7% | 82.6% | 70.0% | 62.6% | 76.1% | 73.1% | 77.0% |
|   Eyes | 50.0% | 50.0% | 50.8% | 50.2% | 50.0% | 50.0% | 50.0% |
|   Eyebrows | 81.9% | 65.0% | 58.5% | 49.8% | 50.0% | 62.0% | 49.5% |
|   Nose | 71.2% | 66.8% | 67.4% | 49.8% | 68.6% | 70.7% | 66.8% |
|   Face contours | 93.8% | 75.3% | 73.2% | 67.3% | 71.2% | 73.1% | 67.9% |
|   Composite face | 97.2% | 85.3% | 77.3% | 75.4% | 80.1% | 82.2% | 82.7% |

**Table 4. Effect of Sexual Dimorphism at the Aggregate Level**

| Context | Linear function | | Quadratic function | | |
|---|---|---|---|---|---|
| | $a_1$ | adj $R^2$ | $b_1$ | $b_2$ | adj $R^2$ |
| Male faces (effect of feminization) | | | | | |
|   Approachability | 0.117*** | 0.753 | 0.255*** | -0.298** | 0.830 |
|   Trustworthiness | 0.065*** | 0.766 | 0.133*** | -0.144* | 0.821 |
|   Competence | 0.016 | 0.082 | 0.033 | -0.035 | 0.039 |
|   Starbucks persona | 0.131*** | 0.793 | 0.261*** | -0.279* | 0.848 |
|   Starbucks customer | 0.119*** | 0.606 | 0.282** | -0.349* | 0.681 |
|   Lab meat customer | 0.065*** | 0.836 | 0.099** | -0.071 | 0.845 |
| Female faces (effect of masculinization) | | | | | |
|   Approachability | -0.139*** | 0.811 | -0.297*** | 0.341** | 0.895 |
|   Trustworthiness | -0.128*** | 0.835 | -0.259*** | 0.283** | 0.905 |
|   Competence | -0.107*** | 0.838 | -0.204*** | 0.210** | 0.891 |
|   Starbucks persona | -0.184*** | 0.831 | -0.405*** | 0.477*** | 0.930 |
|   Starbucks customer | -0.210*** | 0.815 | -0.467*** | 0.556*** | 0.915 |
|   Lab meat customer | -0.112*** | 0.834 | -0.248*** | 0.295*** | 0.937 |

* $p<0.05$, ** $p<0.01$, *** $p<0.001$

**Table 5. Effect of Sexual Dimorphism on Perceptions of Two Individual Evaluators**

| Context | RA | Linear function | | Quadratic function | | |
|---|---|---|---|---|---|---|
| | | $a_1$ | adj $R^2$ | $b_1$ | $b_2$ | adj $R^2$ |
| Male faces (effect of feminization) | | | | | | |
| Approachability | RA1 | 0.049*** | 0.803 | 0.086*** | -0.079 | 0.830 |
| | RA2 | 0.020** | 0.342 | 0.013 | 0.014 | 0.307 |
| Trustworthiness | RA1 | 0.011*** | 0.582 | 0.010 | 0.003 | 0.558 |
| | RA2 | 0.112*** | 0.814 | 0.237*** | -0.267** | 0.889 |
| Competence | RA1 | -0.019*** | 0.494 | -0.051** | 0.069* | 0.589 |
| | RA2 | 0.017* | 0.224 | 0.036 | -0.040 | 0.204 |
| Starbucks persona | RA1 | 0.094*** | 0.826 | 0.188*** | -0.202** | 0.886 |
| | RA2 | 0.042*** | 0.615 | 0.091** | -0.107 | 0.669 |
| Starbucks customer | RA1 | 0.024** | 0.399 | 0.008 | 0.033 | 0.379 |
| | RA2 | 0.122*** | 0.809 | 0.248*** | -0.270** | 0.871 |
| Lab meat customer | RA1 | 0.008 | -0.007 | -0.030 | 0.082 | 0.012 |
| | RA2 | 0.016* | 0.162 | 0.017 | -0.004 | 0.113 |
| Female faces (effect of masculinization) | | | | | | |
| Approachability | RA1 | -0.061*** | 0.710 | -0.135*** | 0.160* | 0.788 |
| | RA2 | -0.016** | 0.465 | -0.015 | -0.003 | 0.433 |
| Trustworthiness | RA1 | -0.017*** | 0.599 | -0.030* | 0.028 | 0.607 |
| | RA2 | -0.117*** | 0.728 | -0.262*** | 0.312** | 0.814 |
| Competence | RA1 | 0.021*** | 0.619 | 0.046** | -0.055* | 0.685 |
| | RA2 | -0.063*** | 0.817 | -0.131*** | 0.148** | 0.894 |
| Starbucks persona | RA1 | -0.086*** | 0.712 | -0.199*** | 0.244** | 0.808 |
| | RA2 | -0.089*** | 0.585 | -0.210** | 0.261* | 0.662 |
| Starbucks customer | RA1 | -0.031*** | 0.555 | -0.068** | 0.079 | 0.600 |
| | RA2 | -0.138*** | 0.752 | -0.314*** | 0.379** | 0.848 |
| Lab meat customer | RA1 | -0.002 | -0.041 | -0.006 | 0.010 | -0.096 |
| | RA2 | -0.051*** | 0.713 | -0.095** | 0.096 | 0.747 |

* $p<0.05$, ** $p<0.01$, *** $p<0.001$

**Table 6. Effect of Sexual Dimorphism Across Different Faces**

| Sign of $a_1$ | Approachability | | Trustworthiness | | Competence | | Starbucks persona | | Starbucks customer | | Lab meat customer | |
|---|---|---|---|---|---|---|---|---|---|---|---|---|
| | Count | % | Count | % | Count | % | Count | % | Count | % | Count | % |
| Male faces (effect of feminization) | | | | | | | | | | | | |
|   Negative | 11 | 3.3 | 13 | 3.9 | 56 | 16.8 | 4 | 1.2 | 2 | 0.6 | 31 | 9.3 |
|   Positive | 301 | 90.4 | 297 | 89.2 | 199 | 59.8 | 307 | 92.2 | 327 | 98.2 | 265 | 79.6 |
|   Non-significant | 21 | 6.3 | 23 | 6.9 | 78 | 23.4 | 22 | 6.6 | 4 | 1.2 | 37 | 11.1 |
| Female faces (effect of masculinization) | | | | | | | | | | | | |
|   Negative | 644 | 89.8 | 673 | 93.9 | 544 | 75.9 | 688 | 96.0 | 704 | 98.2 | 631 | 88.0 |
|   Positive | 17 | 2.4 | 18 | 2.5 | 65 | 9.1 | 7 | 1.0 | 6 | 0.8 | 37 | 5.2 |
|   Non-significant | 56 | 7.8 | 26 | 3.6 | 108 | 15.1 | 22 | 3.1 | 7 | 1.0 | 49 | 6.8 |

**Figure 1. Structure of the TMUD**

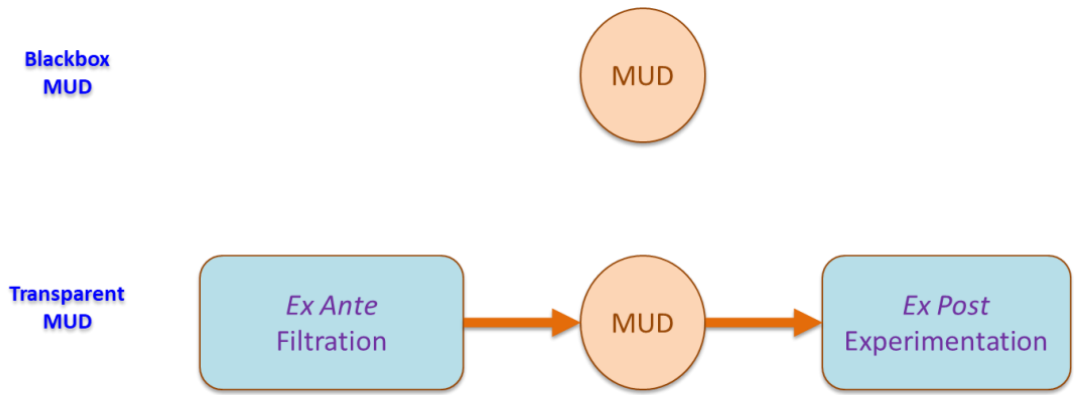

**Figure 2. Structure of MAD and MUD**

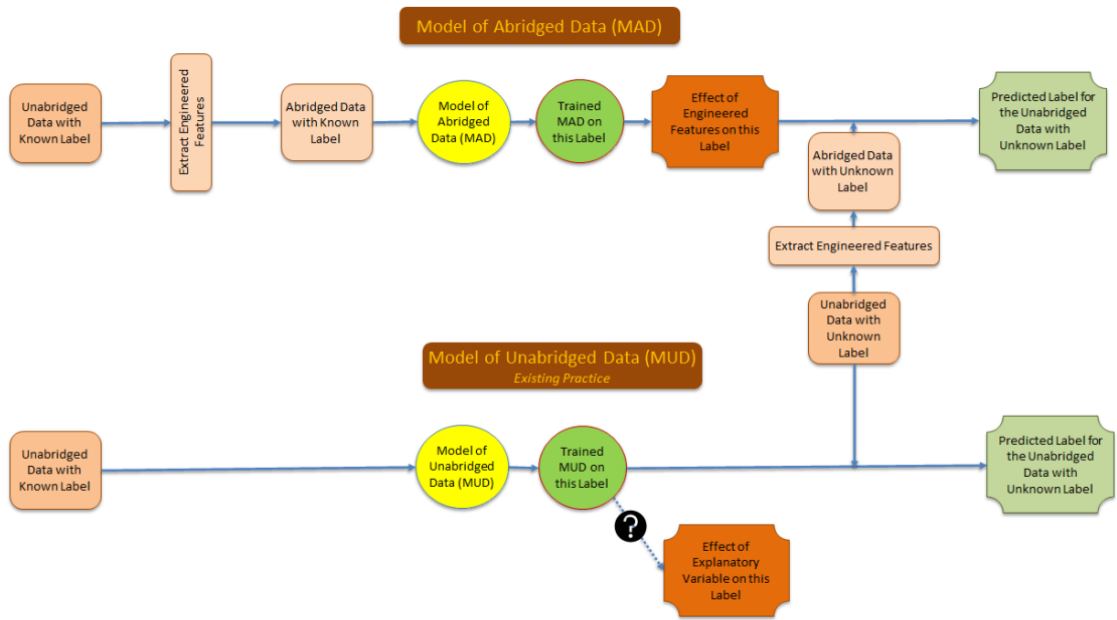

**Figure 3. Biological MUD and Inference Rules**

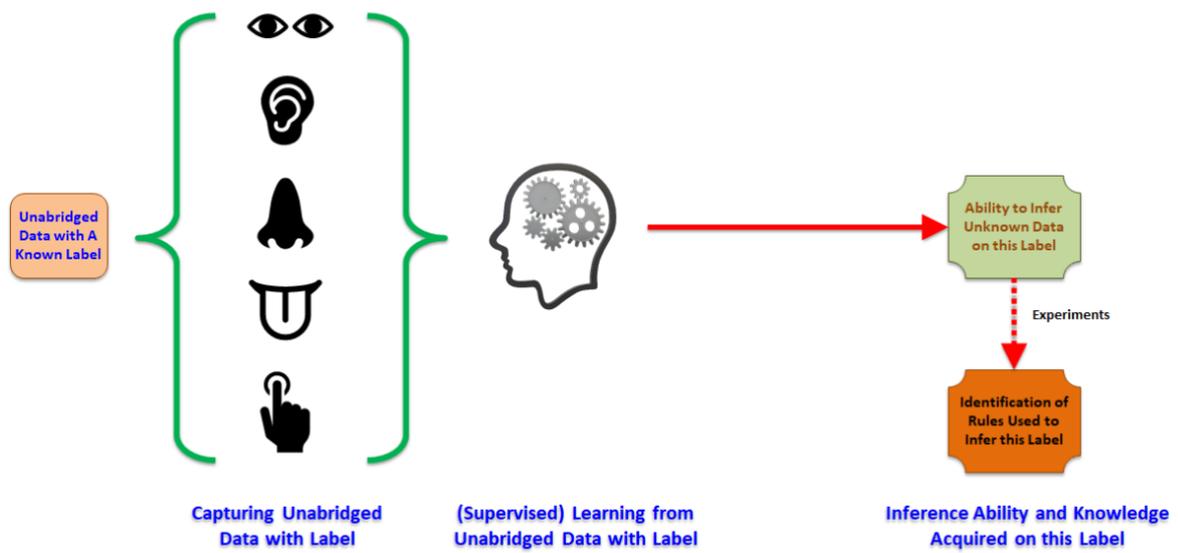

**Figure 4. The *ex post* Experimentation Module of TMUD**

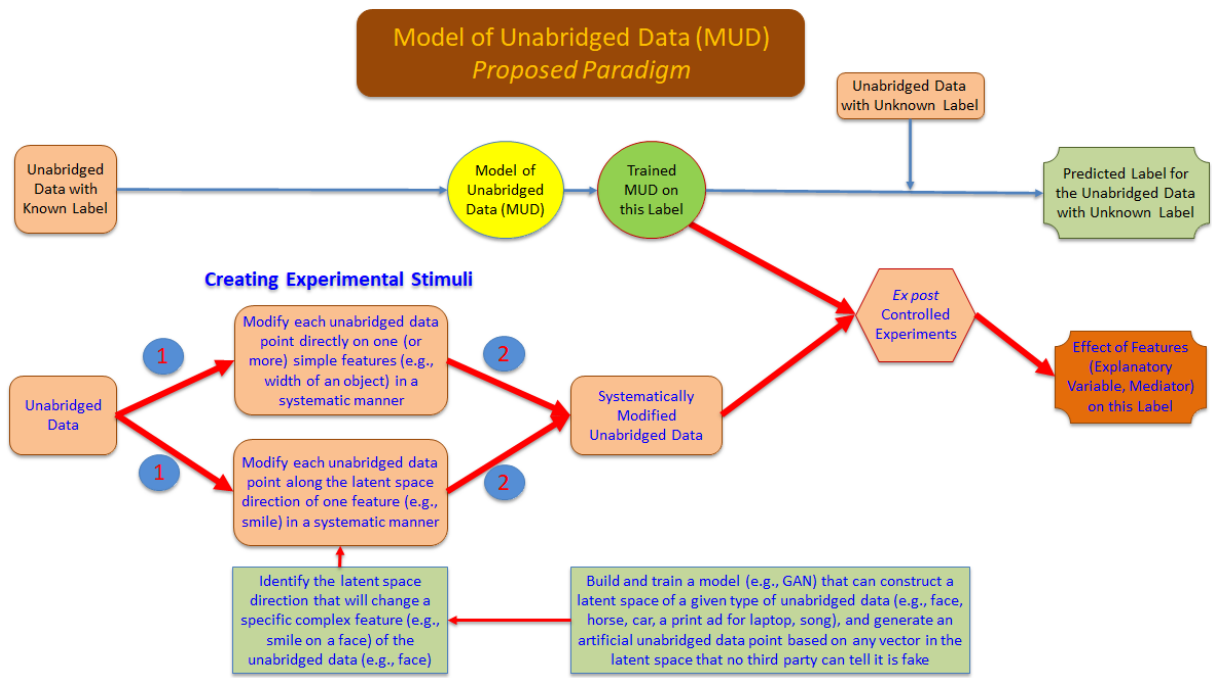

**Figure 5. An Example of Facial Component Extraction**

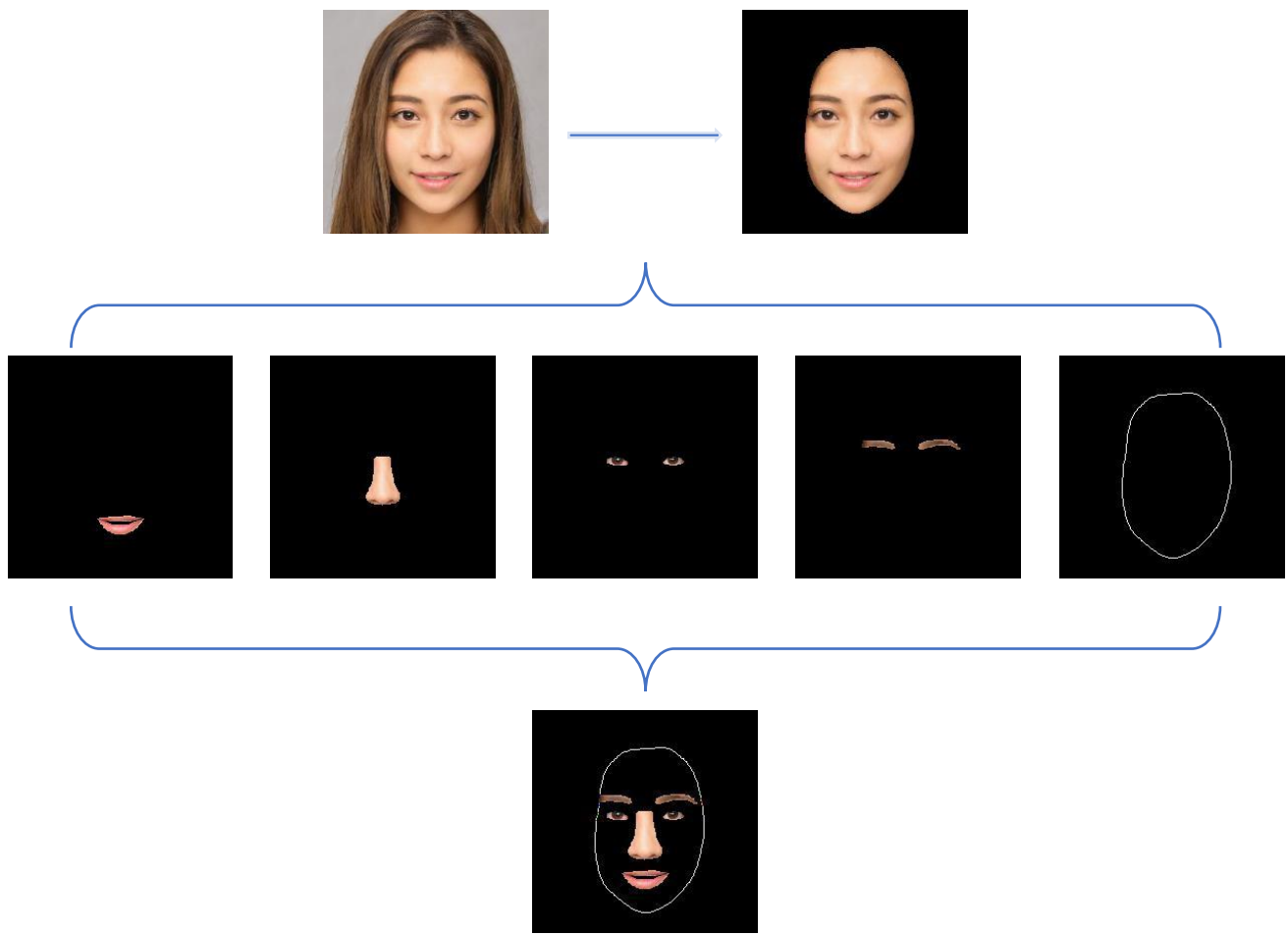

**Figure 6. Two Examples of Face Editing on Sexual Dimorphism**

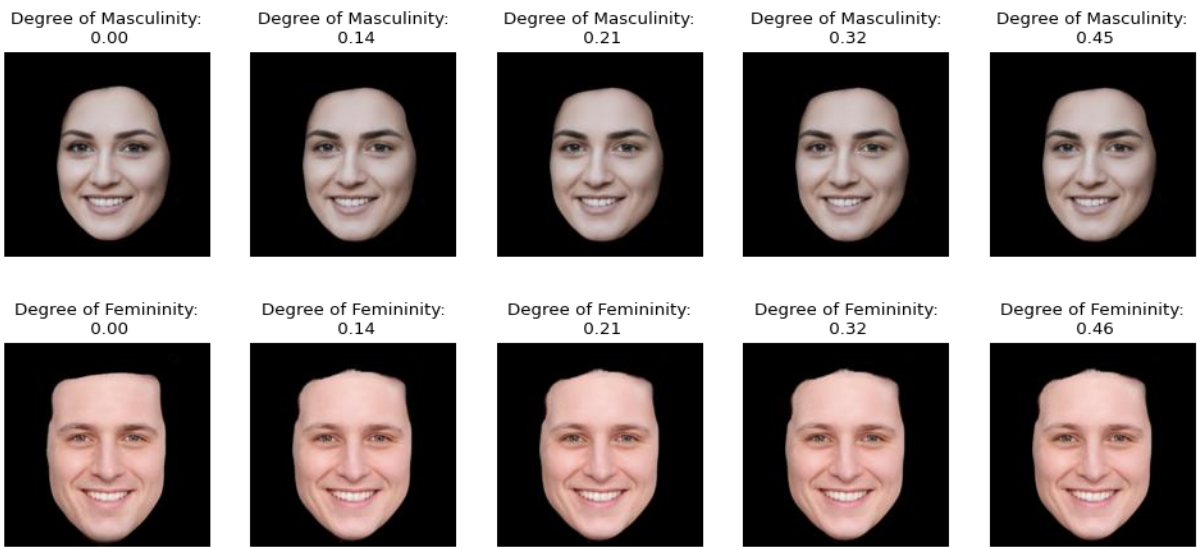

# References

Aslam, S. (2021). Instagram by the numbers: Stats, demographics & fun facts. Retrieved February 19,2021, https://www.omnicoreagency.com/instagram-statistics/.

Babich, N. (2017). Putting personas to work in UX design: what they are and why they're important. *V Adobe Blog*. Retrieved February 19,2021, https://blog.adobe.com/en/publish/2017/09/29/putting-personas-to-work-in-ux-design-what-they-are-and-why-theyre-important.html#gs.1mlmp1.

Bau, D., Zhou, B., Khosla, A., Oliva, A., & Torralba, A. (2017). Network dissection: Quantifying interpretability of deep visual representations. In *Proceedings of the IEEE conference on computer vision and pattern recognition* (pp. 6541-6549).

Behrmann, J., Etmann, C., Boskamp, T., Casadonte, R., Kriegsmann, J., & Maaβ, P. (2018). Deep learning for tumor classification in imaging mass spectrometry. *Bioinformatics*, *34*(7), 1215-1223.

Bengio, Y. (2012). Deep learning of representations for unsupervised and transfer learning. In *Proceedings of ICML workshop on unsupervised and transfer learning* (pp. 17-36). JMLR Workshop and Conference Proceedings.

Broverman, I. K., Vogel, S. R., Broverman, D. M., Clarkson, F. E., & Rosenkrantz, P. S. (1972). Sex-Role Stereotypes: A Current Appraisal 1. *Journal of Social issues*, *28*(2), 59-78.

Burnap, A., Hauser, J. R., & Timoshenko, A. (2020). Design and evaluation of product aesthetics: A human-machine hybrid approach. MIT Sloan Working Paper 5814-19. Cambridge, MA: MIT Sloan School of Management.

Carion, N., Massa, F., Synnaeve, G., Usunier, N., Kirillov, A., & Zagoruyko, S. (2020). End-to-end object detection with transformers. In *European Conference on Computer Vision* (pp. 213-229). Springer, Cham.

Chintagunta, P., Hanssens, D. M., & Hauser, J. R. (2016). Marketing Science and Big Data. *Marketing Science*, *35*(3), 341-342.

Cordts, M., Omran, M., Ramos, S., Rehfeld, T., Enzweiler, M., Benenson, R., ... & Schiele, B. (2016). The cityscapes dataset for semantic urban scene understanding. In *Proceedings of the IEEE conference on computer vision and pattern recognition* (pp. 3213-3223).

Cuddy, A. J., Fiske, S. T., Kwan, V. S., Glick, P., Demoulin, S., Leyens, J. P., ... & Ziegler, R. (2009). Stereotype content model across cultures: Towards universal similarities and some differences. *British Journal of Social Psychology*, *48*(1), 1-33.

DeBruine, L. M., Jones, B. C., Little, A. C., Boothroyd, L. G., Perrett, D. I., Penton-Voak, I. S., ... & Tiddeman, B. P. (2006). Correlated preferences for facial masculinity and ideal or actual partner's masculinity. *Proceedings of the Royal Society B: Biological Sciences*, *273*(1592), 1355-1360.

Dew, R., Ansari, A., & Toubia, O. (2019). Letting logos speak: Leveraging multiview representation learning for data-driven logo design. *Available at SSRN 3406857*.

Dzyabura, D., & Peres, R. (2021), Visual Elicitation of Brand Perception, *Journal of Marketing.*Forthcoming.

Dzyabura, Daria, Siham El Kihal, John R. Hauser, and Marat Ibragimov(2021). Leveraging the power of images in predicting product return rates. MIT Sloan Working Paper 5856-19. Cambridge, MA: MIT Sloan School of Management.

Fink, B., Neave, N., & Seydel, H. (2007). Male facial appearance signals physical strength to women. *American Journal of Human Biology*, *19*(1), 82-87.

Forbes Agency Council (2018) .How To Craft The Ideal User Persona For Your Brand. In Forbes. Retrieved on February 19  2021, https://www.forbes.com/sites/forbesagencycouncil/2018/03/14/how-to-craft-the-ideal-user-persona-for-your-brand/?sh=d6cd7436eadf .

Gabel, S., & Timoshenko, A. (2021). Product Choice with Large Assortments: A Scalable Deep-Learning Model. *Management Science*.

Gabel, S., Guhl, D. and Klapper, D., (2019). P2V-MAP: Mapping Market Structures for Large Retail Assortments. *Journal of Marketing Research*, 56(4), pp.557-580.

Gangestad, S. W., & Scheyd, G. J. (2005). The evolution of human physical attractiveness. *Annu. Rev. Anthropol.*, *34*, 523-548.

Gorn, G. J., Jiang, Y., & Johar, G. V. (2008). Babyfaces, trait inferences, and company evaluations in a public relations crisis. *Journal of Consumer Research*, *35*(1), 36-49.

Guan, Y., Tan, Y., Wei, Q., & Chen, G. (2020). Information or Distortion? The Effect of Customer Generated Images on Product Rating Dynamics. *Available at SSRN 3633590*.

Hartmann, J., Heitmann, M., Schamp, C., & Netzer, O. (2021). The power of brand selfies in consumer-generated brand images. *Columbia Business School Research Paper.* Forthcoming.

Hesamian, M. H., Jia, W., He, X., & Kennedy, P. (2019). Deep learning techniques for medical image segmentation: achievements and challenges. *Journal of digital imaging*, *32*(4), 582-596.

Hess, U., Adams Jr, R. B., & Kleck, R. E. (2004). Facial appearance, gender, and emotion expression. *Emotion*, *4*(4), 378.

Holzwarth, M., Janiszewski, C., & Neumann, M. M. (2006). The influence of avatars on online consumer shopping behavior. *Journal of marketing*, *70*(4), 19-36.

Hu, M., Dang, C., & Chintagunta, P. K. (2019). Search and learning at a daily deals website. *Marketing Science*, *38*(4), 609-642.

Kahle, L. R., & Homer, P. M. (1985). Physical attractiveness of the celebrity endorser: A social adaptation perspective. *Journal of consumer research*, *11*(4), 954-961.

Karras, T., Aila, T., Laine, S., & Lehtinen, J. (2017). Progressive growing of gans for improved quality, stability, and variation. *arXiv preprint arXiv:1710.10196*.

Karras, T., Laine, S., & Aila, T. (2019). A style-based generator architecture for generative adversarial networks. In *Proceedings of the IEEE/CVF Conference on Computer Vision and Pattern Recognition* (pp. 4401-4410).

Karras, T., Laine, S., Aittala, M., Hellsten, J., Lehtinen, J., & Aila, T. (2020). Analyzing and improving the image quality of stylegan. In *Proceedings of the IEEE/CVF Conference on Computer Vision and Pattern Recognition* (pp. 8110-8119).

Krizhevsky, A., Sutskever, I., & Hinton, G. E. (2012). Imagenet classification with deep convolutional neural networks. *Advances in neural information processing systems*, *25*, 1097-1105.

Lapuschkin, S., Binder, A., Muller, K. R., & Samek, W. (2017). Understanding and comparing deep neural networks for age and gender classification. In *Proceedings of the IEEE International Conference on Computer Vision Workshops* (pp. 1629-1638).

LeCun, Y., Bengio, Y., & Hinton, G. (2015). Deep learning. *nature*, *521*(7553), 436-444.

Lester, J. C., Converse, S. A., Kahler, S. E., Barlow, S. T., Stone, B. A., & Bhogal, R. S. (1997). The persona effect: affective impact of animated pedagogical agents. In *Proceedings of the ACM SIGCHI Conference on Human factors in computing systems* (pp. 359-366).

Li, H., Simchi-Levi, D., Wu, M. X., & Zhu, W. (2019). Estimating and Exploiting the Impact of Photo Layout in the Sharing Economy. *Available at SSRN 3470877*.

Li, Y., & Xie, Y. (2020). Is a picture worth a thousand words? An empirical study of image content and social media engagement. *Journal of Marketing Research, 57*(1), 1-19.

Liu, L., Dzyabura, D., & Mizik, N. (2020). Visual listening in: Extracting brand image portrayed on social media. *Marketing Science*, *39*(4), 669-686.

Liu, S., Shi, J., Liang, J., & Yang, M. H. (2017a). Face parsing via recurrent propagation. *arXiv preprint arXiv:1708.01936*.

Liu, X., Lee, D., & Srinivasan, K. (2019). Large-scale cross-category analysis of consumer review content on sales conversion leveraging deep learning. *Journal of Marketing Research*, *56*(6), 918-943.

Liu, X., Shi, S. W., Teixeira, T., & Wedel, M. (2018). Video content marketing: The making of clips. *Journal of Marketing*, *82*(4), 86-101.

Liu, Y., Li, K. J., Chen, H., & Balachander, S. (2017b). The effects of products' aesthetic design on demand and marketing-mix effectiveness: The role of segment prototypicality and brand consistency. *Journal of Marketing*, *81*(1), 83-102.

Lu, S., Xiao, L., & Ding, M. (2016). A video-based automated recommender (VAR) system for garments. *Marketing Science*, *35*(3), 484-510.

Lu, S., Yao, D., Chen, X., & Grewal, R. (2020). Do Larger Audiences Generate Greater Revenues under Pay What You Want? Evidence from a Live Streaming Platform. *Marketing Science*. Forthcoming.

Lyons, R. G. (2004). *Understanding digital signal processing, 3/E*. Pearson Education India.

Malik, N., Singh, P. V., & Srinivasan, K. (2019). A Dynamic Analysis of Beauty Premium. *Available at SSRN 3208162*.

McDuff, D., & Berger, J. (2019). Do Facial Expressions Predict Ad Sharing? A Large-Scale Observational Study. *arXiv preprint arXiv:1912.10311*.

Messer, U., & Fausser, S. (2019). Predicting Social Perception from Faces: A Deep Learning Approach. *arXiv preprint arXiv:1907.00217*.

Minaee, S., Boykov, Y. Y., Porikli, F., Plaza, A. J., Kehtarnavaz, N., & Terzopoulos, D. (2021). Image segmentation using deep learning: A survey. *IEEE Transactions on Pattern Analysis and Machine Intelligence*.

Oh, D., Buck, E. A., & Todorov, A. (2019). Revealing hidden gender biases in competence impressions of faces. *Psychological Science*, *30*(1), 65-79.

Olah, C., Cammarata, N., Schubert, L., Goh, G., Petrov, M., & Carter, S. (2020). Zoom in: An introduction to circuits. *Distill*, *5*(3), e00024-001.

Oosterhof, N. N., & Todorov, A. (2008). The functional basis of face evaluation. *Proceedings of the National Academy of Sciences*, *105*(32), 11087-11092.

Oppenheim, A. V. (1999). *Discrete-time signal processing*. Pearson Education India.

Parkhi, O. M., Vedaldi, A., & Zisserman, A. (2015). Deep Face Recognition. Proceedings of the British Machine Vision Conference (BMVC), pages 41.1-41.12. BMVA Press, September 2015.

Peng, L., Cui, G., Chung, Y., & Zheng, W. (2020). The faces of success: Beauty and ugliness premiums in e-commerce platforms. *Journal of Marketing*, *84*(4), 67-85.

Pruitt, J., & Adlin, T. (2010). *The persona lifecycle: keeping people in mind throughout product design*. Elsevier.

Radford, A., Metz, L., & Chintala, S. (2015). Unsupervised representation learning with deep convolutional generative adversarial networks. *arXiv preprint arXiv:1511.06434*.

Raghu, M., & Schmidt, E. (2020). A survey of deep learning for scientific discovery. *arXiv preprint arXiv:2003.11755*.

Rizkallah, J. (2017). The big (unstructured) data problem. *Forbes*(June 5), https://www.forbes.com/sites/forbestechcouncil/2017/06/05/the-big-unstructured-data-problem/?sh=da78e05493a3.

Rule, N. O., & Ambady, N. (2008). The face of success: Inferences from chief executive officers' appearance predict company profits. *Psychological science*, *19*(2), 109-111.
55

Selvaraju, R. R., Cogswell, M., Das, A., Vedantam, R., Parikh, D., & Batra, D. (2017). Grad-cam: Visual explanations from deep networks via gradient-based localization. In *Proceedings of the IEEE international conference on computer vision* (pp. 618-626).

Shen S.(2018). What makes effeminate men popular in China. *Global Times*(September 12), https://www.globaltimes.cn/content/1119350.shtml.

Shen, Y., Gu, J., Tang, X., & Zhou, B. (2020). Interpreting the latent space of gans for semantic face editing. In *Proceedings of the IEEE/CVF Conference on Computer Vision and Pattern Recognition* (pp. 9243-9252).

Shin, D., He, S., Lee, G. M., WHINSTON, A. B., Centintas, S., & Lee, K. C. (2020). Enhancing Social Media Analysis with Visual Data Analytics: A Deep Learning Approach. *Management Information Systems Quarterly*, 44(4), 1459-1492.

Shorten, C., & Khoshgoftaar, T. M. (2019). A survey on image data augmentation for deep learning. *Journal of Big Data*, *6*(1), 1-48.

Simonyan, K., Vedaldi, A., & Zisserman, A. (2013). Deep inside convolutional networks: Visualising image classification models and saliency maps. *arXiv preprint arXiv:1312.6034*.

Smith, B. M., Zhang, L., Brandt, J., Lin, Z., & Yang, J. (2013). Exemplar-based face parsing. In *Proceedings of the IEEE conference on computer vision and pattern recognition* (pp. 3484-3491).

So, C., & Joo, J. (2017). Does a persona improve creativity?. *The Design Journal*, *20*(4), 459-475.

Sofer, C., Dotsch, R., Oikawa, M., Oikawa, H., Wigboldus, D. H., & Todorov, A. (2017). For your local eyes only: Culture-specific face typicality influences perceptions of trustworthiness. *Perception*, *46*(8), 914-928.

Srinivasan, V., & Park, C. S. (1997). Surprising robustness of the self-explicated approach to customer preference structure measurement. *Journal of Marketing Research*, *34*(2), 286-291.

Sutherland, C. A., Oldmeadow, J. A., Santos, I. M., Towler, J., Burt, D. M., & Young, A. W. (2013). Social inferences from faces: Ambient images generate a three-dimensional model. *Cognition*, *127*(1), 105-118.

Teixeira, T., Picard, R., & El Kaliouby, R. (2014). Why, when, and how much to entertain consumers in advertisements? A web-based facial tracking field study. *Marketing Science*, *33*(6), 809-827.

Teixeira, T., Wedel, M., & Pieters, R. (2012). Emotion-induced engagement in internet video advertisements. *Journal of marketing research*, *49*(2), 144-159.

Tellis, G. J., MacInnis, D. J., Tirunillai, S., & Zhang, Y. (2019). What drives virality (sharing) of online digital content? The critical role of information, emotion, and brand prominence. *Journal of Marketing*, *83*(4), 1-20.

Timoshenko, A., & Hauser, J. R. (2019). Identifying customer needs from user-generated content. *Marketing Science*, *38*(1), 1-20.

Tkachenko, Y., & Jedidi, K. (2020). What Personal Information Can a Consumer Facial Image Reveal? Implications for Marketing ROI and Consumer Privacy. *Available at SSRN 3616470*.

Todorov, A., Dotsch, R., Porter, J. M., Oosterhof, N. N., & Falvello, V. B. (2013). Validation of data-driven computational models of social perception of faces. *Emotion*, *13*(4), 724.

Todorov, A., Said, C. P., Engell, A. D., & Oosterhof, N. N. (2008). Understanding evaluation of faces on social dimensions. *Trends in cognitive sciences*, *12*(12), 455-460.

Troncoso, I., & Luo, L. (2020). Look the Part? The Role of Profile Pictures in Online Labor Markets. *Available at SSRN 3709554*.

Urban, G., Timoshenko, A., Dhillon, P., & Hauser, J. R. (2020). Is Deep Learning a Game Changer for Marketing Analytics?. *MIT Sloan Management Review, 61*(2), 70-76.

van der Lans, R., Pieters, R., & Wedel, M. (2021). Online Advertising Suppresses Visual Competition during Planned Purchases. *Journal of Consumer Research*.

Vernon, R. J., Sutherland, C. A., Young, A. W., & Hartley, T. (2014). Modeling first impressions from highly variable facial images. *Proceedings of the National Academy of Sciences*, *111*(32), E3353-E3361.

Walker, M., & Wänke, M. (2017). Caring or daring? Exploring the impact of facial masculinity/femininity and gender category information on first impressions. *PloS one*, *12*(10), e0181306.

Wolf, N. (2013). *The beauty myth: How images of beauty are used against women*. Random House.

Xia, F., Chatterjee, R., & May, J. H. (2019). Using conditional restricted Boltzmann machines to model complex consumer shopping patterns. *Marketing Science*, *38*(4), 711-727.

Xiao, L., & Ding, M. (2014). Just the faces: Exploring the effects of facial features in print advertising. *Marketing Science*, *33*(3), 338-352.

Xiao, L., Kim, H. J., & Ding, M. (2013). An Introduction to Audio and Visual Research and Applications in Marketing. *Review of Marketing Research, 10,* 213-253.

Yu, C., Wang, J., Peng, C., Gao, C., Yu, G., & Sang, N. (2018). Bisenet: Bilateral segmentation network for real-time semantic segmentation. In *Proceedings of the European conference on computer vision (ECCV)* (pp. 325-341).

Zhan, J., Liu, M., Garrod, O. G., Daube, C., Ince, R. A., Jack, R. E., & Schyns, P. G. (2021). Modeling individual preferences reveals that face beauty is not universally perceived across cultures. *Current Biology*.

Zhang, M. and Luo, L. (2018). Can User-Posted Photos Serve as a Leading Indicator of Restaurant Survival? Evidence from Yelp. *Available at SSRN 3108288*.

Zhang, Q., Wang, W., & Chen, Y. (2020). Frontiers: In-Consumption Social Listening with Moment-to-Moment Unstructured Data: The Case of Movie Appreciation and Live Comments. *Marketing Science*, *39*(2), 285-295.

Zhang, S., Lee, D. D., Singh, P. V., & Srinivasan, K. (2017). How much is an image worth? Airbnb property demand estimation leveraging large scale image analytics. *Available at SSRN 2976021*.

Zhang, S., Mehta, N., Singh, P. V., & Srinivasan, K. (2021). Can an AI Algorithm Mitigate Racial Economic Inequality? An Analysis in the Context of Airbnb. *Marketing Science*, Forthcoming.

Zhang, X., Li, S., Burke, R. R., & Leykin, A. (2014). An examination of social influence on shopper behavior using video tracking data. *Journal of Marketing*, *78*(5), 24-41.

Zheng, T. (2015). Masculinity in crisis: effeminate men, loss of manhood, and the nation-state in postsocialist China. *Etnográfica. Revista do Centro em Rede de Investigação em Antropologia*, *19*(2)), 347-365.

Zhou, Y., Lu, S., & Ding, M. (2020). Contour-as-Face Framework: A Method to Preserve Privacy and Perception. *Journal of Marketing Research*, *57*(4), 617-639.

Zhuang, F., Qi, Z., Duan, K., Xi, D., Zhu, Y., Zhu, H., ... & He, Q. (2020). A comprehensive survey on transfer learning. *Proceedings of the IEEE*, *109*(1), 43-76.
58